%% file: main.tex
\documentclass[letterpaper]{article}
\usepackage[preprint]{aaai2027}
\input{shared_preamble}
\usepackage{tikz}
\usepackage[most]{tcolorbox}
\usepackage{enumitem}
\usetikzlibrary{positioning,arrows.meta,shapes.geometric}

\makeatletter
\patchcmd{\@maketitle}
  {Corresponding author\ifaaai@corrmulti{s}\fi.}
  {Corresponding authors.}{}{}
\makeatother

\newcommand{\projectpagelink}{%
  \leavevmode\pdfstartlink attr{/Border [0 0 0]} user{%
    /Subtype /Link /A << /S /URI /URI (https://kendrick-stein.github.io/InteractiveRewardAgent-OfficialRepo/) >>}%
  \textcolor{blue}{\textbf{Project Page}}%
  \pdfendlink%
}

\definecolor{RABlue}{RGB}{42,132,222}
\definecolor{RABlueBg}{RGB}{237,246,255}
\definecolor{RAGray}{RGB}{241,243,246}
\definecolor{RAGrayLine}{RGB}{216,222,229}
\definecolor{RAMuted}{RGB}{94,107,122}
\definecolor{RAGreen}{RGB}{46,160,84}
\definecolor{RAGreenBg}{RGB}{237,252,242}
\definecolor{RARed}{RGB}{209,48,48}

\newtcbox{\raBadge}{on line, arc=2pt, boxrule=0.4pt,
  colframe=RABlue, colback=white, left=3pt, right=3pt,
  top=1pt, bottom=1pt, boxsep=0pt, fontupper=\scriptsize\sffamily}
\newtcbox{\raGoodBadge}{on line, arc=2pt, boxrule=0.4pt,
  colframe=RAGreen, colback=white, left=3pt, right=3pt,
  top=1pt, bottom=1pt, boxsep=0pt, fontupper=\scriptsize\sffamily\color{RAGreen}}
\newtcbox{\raBadBadge}{on line, arc=2pt, boxrule=0.4pt,
  colframe=RARed, colback=white, left=3pt, right=3pt,
  top=1pt, bottom=1pt, boxsep=0pt, fontupper=\scriptsize\sffamily\color{RARed}}

\newtcolorbox{rainputbox}{enhanced,breakable,colback=RABlueBg,
  colframe=RABlue,boxrule=0.5pt,arc=2pt,left=6pt,right=6pt,
  top=6pt,bottom=6pt,before skip=4pt,after skip=6pt}
\newtcolorbox{ratoolbox}{enhanced,breakable,colback=RAGray,
  colframe=RAGrayLine,boxrule=0.4pt,arc=0pt,borderline west={1.2pt}{0pt}{RAMuted},
  left=7pt,right=5pt,top=5pt,bottom=5pt,before skip=4pt,after skip=6pt}
\newtcolorbox{raguibox}{enhanced,breakable,colback=RABlueBg,
  colframe=RABlue,boxrule=0.5pt,arc=2pt,left=6pt,right=6pt,
  top=6pt,bottom=6pt,before skip=4pt,after skip=6pt}
\newtcolorbox{rafinalbox}{enhanced,breakable,colback=RAGreenBg,
  colframe=RAGreen,boxrule=0.5pt,arc=2pt,left=6pt,right=6pt,
  top=6pt,bottom=6pt,before skip=4pt,after skip=6pt}

\newcommand{\raBoxTitle}[1]{{\ttfamily\footnotesize\bfseries #1\par}}
\newcommand{\raStepHeader}[2]{{\ttfamily\scriptsize\bfseries [#1] #2\par}}
\newcommand{\raCmd}[1]{{\ttfamily\scriptsize\raggedright #1\par}}
\newcommand{\raThought}[1]{{\normalfont\itshape\scriptsize\color{RAMuted}#1\par}}
\newcommand{\raObs}[1]{{\ttfamily\scriptsize\color{RAMuted}#1\par}}
\newcommand{\raFinalText}[1]{{\normalfont\scriptsize #1\par}}
\newcommand{\raGuiStepHeader}[2]{{\ttfamily\small\bfseries [#1] #2\par}}
\newcommand{\raGuiCmd}[1]{{\ttfamily\small\raggedright #1\par}}
\newcommand{\raGuiThought}[1]{{\normalfont\itshape\small\raggedright\color{RAMuted}#1\par}}
\newcommand{\raGuiObs}[1]{{\ttfamily\footnotesize\raggedright\color{RAMuted}#1\par}}

\newcommand{\raComputerPointImage}[4]{%
\begin{tikzpicture}
\node[anchor=south west,inner sep=0] (raimage) at (0,0) {\includegraphics[width=\linewidth]{#1}};
\pgfmathsetmacro{\rax}{#2/1000}
\pgfmathsetmacro{\ray}{1-(#3/1000)}
\begin{scope}[x={(raimage.south east)},y={(raimage.north west)}]
\coordinate (rapoint) at (\rax,\ray);
\node[draw=RARed,fill=white,circle,line width=0.55pt,minimum size=5.2pt,inner sep=0pt] at (rapoint) {};
\node[anchor=south west,xshift=3pt,yshift=3pt,fill=white,fill opacity=0.92,text opacity=1,
  draw=RARed,rounded corners=0.8pt,line width=0.4pt,inner xsep=2pt,inner ysep=0.8pt,
  font=\tiny\ttfamily,text=RARed] at (rapoint) {#4};
\end{scope}
\end{tikzpicture}%
}

\newcommand{\raComputerLabelImage}[2]{%
\begin{tikzpicture}
\node[anchor=south west,inner sep=0] (raimage) at (0,0) {\includegraphics[width=\linewidth]{#1}};
\begin{scope}[x={(raimage.south east)},y={(raimage.north west)}]
\node[anchor=north east,fill=white,fill opacity=0.92,text opacity=1,
  draw=RARed,rounded corners=0.8pt,line width=0.4pt,inner xsep=2pt,inner ysep=0.8pt,
  font=\tiny\ttfamily,text=RARed] at (0.985,0.965) {#2};
\end{scope}
\end{tikzpicture}%
}

\definecolor{ECBlue}{RGB}{42,132,222}
\definecolor{ECBlueBg}{RGB}{237,246,255}
\definecolor{ECGray}{RGB}{241,243,246}
\definecolor{ECGrayLine}{RGB}{216,222,229}
\definecolor{ECMuted}{RGB}{94,107,122}
\definecolor{ECGreen}{RGB}{46,160,84}
\definecolor{ECGreenBg}{RGB}{237,252,242}
\definecolor{ECRed}{RGB}{209,48,48}
\definecolor{ECRedBg}{RGB}{255,245,245}

\newtcbox{\ecBadge}{on line, arc=2pt, boxrule=0.4pt,
  colframe=ECBlue, colback=white, left=3pt, right=3pt,
  top=1pt, bottom=1pt, boxsep=0pt, fontupper=\scriptsize\sffamily}
\newtcbox{\ecGoodBadge}{on line, arc=2pt, boxrule=0.4pt,
  colframe=ECGreen, colback=white, left=3pt, right=3pt,
  top=1pt, bottom=1pt, boxsep=0pt, fontupper=\scriptsize\sffamily\color{ECGreen}}
\newtcbox{\ecBadBadge}{on line, arc=2pt, boxrule=0.4pt,
  colframe=ECRed, colback=white, left=3pt, right=3pt,
  top=1pt, bottom=1pt, boxsep=0pt, fontupper=\scriptsize\sffamily\color{ECRed}}

\newtcolorbox{ecinputbox}{enhanced, breakable, colback=ECBlueBg,
  colframe=ECBlue, boxrule=0.5pt, arc=2pt, left=6pt, right=6pt,
  top=6pt, bottom=6pt, before skip=4pt, after skip=6pt}
\newtcolorbox{ectoolbox}{enhanced, breakable, colback=ECGray,
  colframe=ECGrayLine, boxrule=0.4pt, arc=0pt, borderline west={1.2pt}{0pt}{ECMuted},
  left=7pt, right=5pt, top=5pt, bottom=5pt, before skip=4pt, after skip=6pt}
\newtcolorbox{ecfinalbox}{enhanced, breakable, colback=ECGreenBg,
  colframe=ECGreen, boxrule=0.5pt, arc=2pt, left=6pt, right=6pt,
  top=6pt, bottom=6pt, before skip=4pt, after skip=8pt}

\newcommand{\ecBoxTitle}[1]{{\ttfamily\footnotesize\bfseries #1\par}}
\newcommand{\ecStepHeader}[2]{{\ttfamily\scriptsize\bfseries [#1] #2\par}}
\newcommand{\ecThought}[1]{{\normalfont\itshape\scriptsize\color{ECMuted}#1\par}}
\newcommand{\ecFinalText}[1]{{\normalfont\scriptsize #1\par}}

\title{Interactive Reward Agent: GUI Task Evaluation \\ via Environment-State Verification}
\author{
Chenrui Shi\textsuperscript{\rm 1,2},
Yuwei Wu\textsuperscript{\rm 1,3},
Yang Liu\textsuperscript{\rm 2},
Ruining Feng\textsuperscript{\rm 2,4},\\
Zirui Shang\textsuperscript{\rm 1,2},
Zhi Gao\textsuperscript{\rm 1,2,3}\corresponding,
Lifeng Fan\textsuperscript{\rm 2},
Che Sun\textsuperscript{\rm 3}\corresponding
}
\affiliations{
\textsuperscript{\rm 1}Beijing Key Laboratory of Intelligent Information Technology, School of Computer Science \& Technology,\\
Beijing Institute of Technology, Beijing, China\\
\textsuperscript{\rm 2}State Key Laboratory of General Artificial Intelligence, BIGAI, Beijing, China\\
\textsuperscript{\rm 3}Guangdong Laboratory of Machine Perception and Intelligent Computing, Shenzhen MSU-BIT University, Shenzhen, China\\
\textsuperscript{\rm 4}Tsinghua University, Beijing, China\\[0.2em]
\texttt{\{shichenrui,wuyuwei,shangzirui,gaozhibit\}@bit.edu.cn}\\
\texttt{\{liuyang,fengruining,fanlifeng\}@bigai.ai}, \texttt{sunche@smbu.edu.cn}\\[0.2em]
\projectpagelink
}
\begin{document}

\maketitle

\begin{abstract}
Graphical user interface task evaluation aims to determine whether a GUI agent has successfully completed a user instruction. Automated GUI task evaluation has received increasing attention because the evaluation results can serve as reward signals for both test-time scaling and post-training. However, reliable GUI task evaluation remains challenging because the judgments often require access to environment states, such as system configurations, file data, and application settings, beyond the screenshots of execution trajectories. In this paper, we propose an \textbf{interactive reward agent (IRA)} based on a propose-then-verify framework to acquire and verify evidence from the post-execution environment. Given a task instruction and a GUI environment after the GUI agent execution, IRA first proposes the task completion conditions and then verifies them by invoking system tools, application tools, and GUI tools. This design combines evidence from both visible interfaces and the environment state in an interactive process. We further introduce \textbf{GUI-RewardBench}, a benchmark of 321 GUI task trajectories spanning 10 Ubuntu desktop application categories. Experiments show that IRA achieves 86.9\% accuracy on GUI-RewardBench, outperforming existing evaluator baselines. We further apply IRA to reinforcement learning of GUI agents, achieving a  34.0\% OSWorld success rate, which demonstrates that IRA can provide effective reward signals for training GUI agents.
\end{abstract}

\input{sections/introduction}

\input{sections/related_work}

\input{sections/method}
\input{sections/benchmark}
\input{sections/experiments}
\input{sections/conclusion}

\bibliography{references}

\clearpage

\input{supplementary_sections/appendix}

\input{supplementary_sections/error_case_analysis}

\end{document}

%% file: shared_preamble.tex
\usepackage[hyphens]{url}
\usepackage{graphicx}
\usepackage{natbib}
\usepackage{caption}
\usepackage{booktabs}
\usepackage{multirow}
\usepackage{amsmath,amsfonts}
\usepackage{array}
\usepackage{textcomp}
\usepackage{xcolor}

%% file: sections/introduction.tex
\section{Introduction}
\begin{figure}[t]
\centering
\includegraphics[width=1\linewidth]{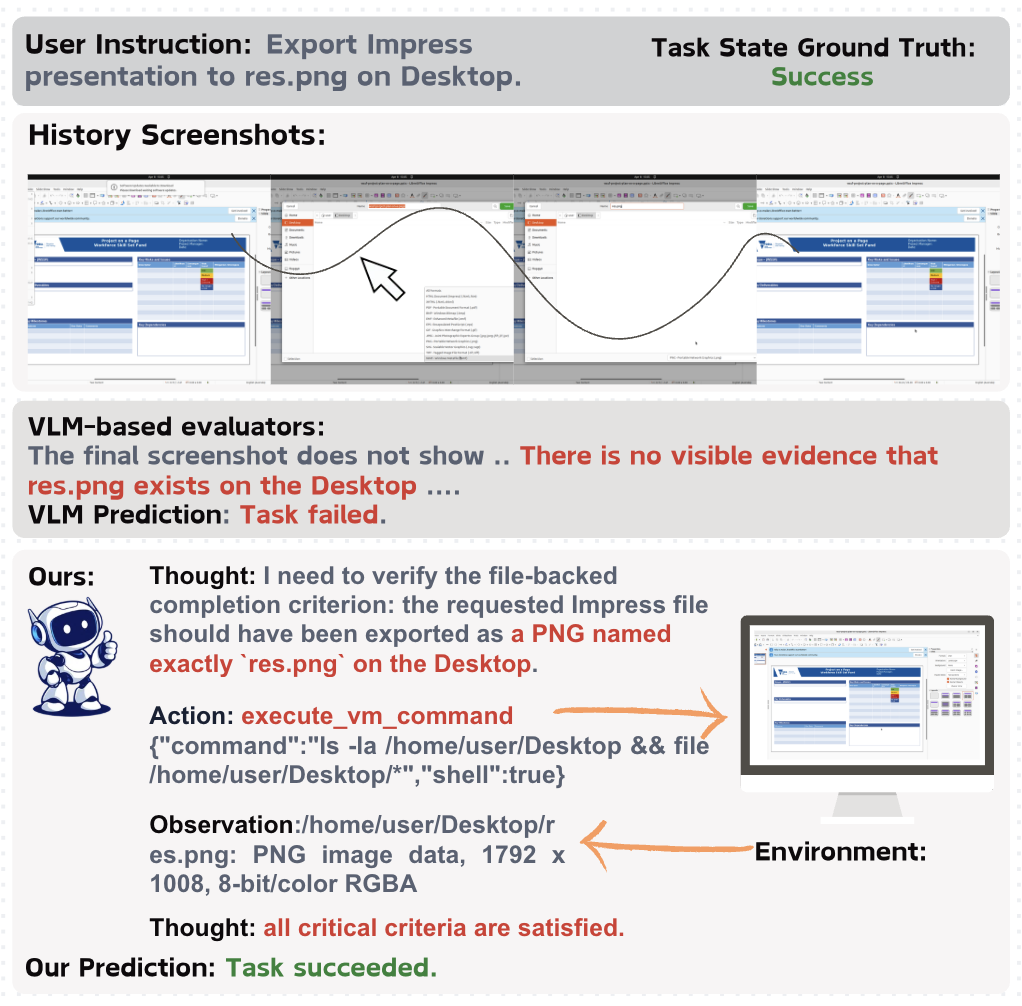}
\caption{Comparison between the interactive reward agent (ours) and a VLM-based evaluator on a task that requires output-file verification.}
\label{fig:example}
\end{figure}

Graphical user interface (GUI) task evaluation aims to determine whether a GUI agent has successfully completed a user instruction. Recently, automated GUI task evaluation has attracted growing attention because reliable automated evaluation is important for both testing and training. For testing, automated evaluation enables test-time scaling~\cite{Lai2025,GonzalezPumariega2025}. For training, evaluation signals can serve directly as rewards for post-training~\cite{Xu2024a,Beyond2026,jiang2026recovering}.  

Existing methods provide a strong foundation for automated evaluation of GUI tasks. 
Script-based evaluators used in existing GUI benchmarks can inspect files, system configurations, application states, and environment variables to determine task success~\cite{Zhou2023a,Bonatti2024,Sun2024,osworld,agentstudio,worldbits}. At the same time, recent progress in vision-language models (VLMs) has enabled more general evaluation methods that judge task completion from screenshots of execution
trajectories~\cite{Hong2023,Lu2024,Zhang2025a,Wu2024,Qin2025,Hu2025}. These VLM-based evaluators reduce the need for per-task manually annotated scripts and make scalable evaluation practical. 
However, task completion may also be reflected in environment-state evidence, such as system configurations, file data, and application settings, beyond what screenshots reveal~\cite{Zheng2024a,Wang2024c,Ye2026,Harnessing2024,Navigating2024,Liu2024b}. As Fig.~\ref{fig:example} shows, exporting a LibreOffice Impress presentation as \texttt{res.png} to the Desktop creates an output file whose existence and content must be verified outside the application window. This example illustrates that reliable GUI task evaluation requires not only visual understanding but also access to environment states.

This observation motivates us to formulate GUI task evaluation as an interaction process between the evaluator and the environment, in which the evaluator iteratively collects evidence before reaching a final judgment. In doing so, we must address a central challenge for interactive evaluation of GUI task completion. 
The required evidence is often distributed across heterogeneous sources, including system configurations, file data, and application settings, whose locations and relevance cannot be determined in advance. Different tasks may require different verification strategies, and a fixed checking procedure may fail when interfaces change, applications behave unexpectedly, or relevant evidence is only partially observable. As a result, reliable GUI evaluation requires not only identifying what conditions should be verified, but also dynamically determining how the corresponding evidence should be obtained from the environment and validated.

In this paper, we propose an interactive reward agent (IRA) built on a propose-then-verify framework and equipped with tools that interact with the environment. Given a task instruction and the post-execution GUI environment, IRA first generates explicit task completion conditions and then verifies them by invoking these tools. Concretely, IRA uses a VLM to propose task-specific completion conditions, rather than directly judging task success. It then verifies these conditions by collecting evidence from the actual environment through system tools, application tools, and GUI tools. This framework retains the scalability of VLM-based evaluators across diverse tasks while enabling evidence-grounded checks similar to those used by script-based evaluators~\cite{wei2026opencomputer,Chen2024c}.

We further introduce \textbf{GUI-RewardBench}, a benchmark of 321 task trajectories spanning 10 Ubuntu desktop application categories, to evaluate how accurately evaluators judge GUI task completion in practical desktop environments. GUI-RewardBench covers visible-state tasks that can be verified from screenshots, hidden-state tasks that require inspecting application configurations or system settings, and artifact-verification tasks that require checking generated files. The benchmark includes applications with substantially different verification characteristics.

Experiments on GUI-RewardBench show that IRA provides strong and consistent evaluation performance. IRA achieves 86.9\% overall accuracy, outperforming existing VLM-based evaluators. The improvement is especially clear in categories where completion evidence frequently resides in hidden environment states, such as VLC, Thunderbird, and multi-application workflows. The human--IRA agreement analysis on trajectories produced from 100 automatically generated tasks indicates that IRA's judgments are strongly aligned with those of human annotators. These results support the value of the propose-then-verify framework for scalable GUI task evaluation. We further use IRA to provide rewards for reinforcement learning of GUI agents. The resulting experiment suggests that IRA can offer training signals comparable to ground-truth evaluation scripts.

Our contributions are summarized as follows:
\begin{itemize}
    \item We propose an interactive reward agent (IRA), an interactive GUI task evaluator based on a propose-then-verify framework that combines the scalability of VLM-based evaluators with the evidence-grounded verification of script-based evaluators.
    \item We introduce GUI-RewardBench, a benchmark of 321 GUI task trajectories whose final states are stable under replay across 10 Ubuntu desktop application categories.
    \item Experimental results show that IRA achieves 86.9\% accuracy on GUI-RewardBench, outperforming existing evaluator baselines.
\end{itemize}

%% file: sections/related_work.tex
\section{Related Work}
\label{sec:relatedwork}

\noindent\textbf{GUI Agents.} GUI agents have progressed from web agents based on structured DOM or HTML observations~\cite{worldbits,mind2web} to VLM-based systems that operate across web, mobile, and desktop environments. Recent work has improved visual grounding using screenshots or parsed interface elements~\cite{Hong2023,Harnessing2024,Lu2024,Wu2024,zhang2025tonguiinternetscaletrajectoriesmultimodal}, extended GUI agents to end-to-end computer use~\cite{Qin2025,Wang2025b,Agashe2025,Lai2025}, and enhanced long-horizon execution through reinforcement learning, verifier feedback, adaptive data curation, and task-specific knowledge~\cite{Xu2024a,zhang2025improving,ding2023multiagent,zhang2025enhancing,wang2025online,liu2026enhancing,xie2026guide}. As GUI agents become increasingly capable, reliably evaluating whether they have actually satisfied a user instruction becomes correspondingly important. However, many recent training and evaluation pipelines infer task success primarily from screenshot trajectories, which provide only a partial view of the environment~\cite{digirl,wang2025distrl,qi2025webrl}. They may miss changes to files, configurations, or backend states, and the displayed interface may not reflect the current system state.
Consequently, a visually plausible trajectory or a final screenshot does not necessarily confirm task completion. IRA addresses this limitation by deriving completion conditions from the instruction and actively verifying them against the final environment state.

\noindent\textbf{GUI Benchmarks.} Existing GUI benchmarks cover a broad range of capabilities, including static interface grounding~\cite{Li2025a}, GUI knowledge~\cite{shi2025gui}, cross-application mobile navigation~\cite{Comprehensive2024}, realistic web interaction~\cite{Zhou2023a,mind2web,chezelles2024browsergym}, and executable mobile, desktop, and computer-use environments~\cite{wu2026mobilegym,wang2026cua,osworld,Bonatti2024,agentstudio,Process2026}. Cross-platform suites further evaluate capabilities ranging from interface understanding and grounding to task automation and collaboration~\cite{Wang2024c}. Despite their diversity, these benchmarks primarily ask whether a GUI agent can execute a task successfully. GUI-RewardBench studies a different problem. It asks whether an evaluator can correctly determine task completion. This shifts the evaluation target from action generation to completion verification, requiring evaluators to reason over heterogeneous evidence such as screenshots, files, configurations, generated artifacts, and application states. GUI-RewardBench therefore complements existing agent-capability benchmarks by directly evaluating the reliability of reward evaluators.

\noindent\textbf{GUI Task Evaluation.} Existing GUI task evaluation methods can be broadly categorized as script-based, VLM-based, and agent-based. Script-based evaluators directly inspect system or application states and can provide precise task-specific checks, but they require substantial manual labor for annotating each task~\cite{Zhou2023a,Sun2024,osworld,agentstudio,worldbits}. VLM-based evaluators are easier to scale across tasks, but typically judge completion from screenshots or trajectories and are therefore limited to visible evidence~\cite{wang2025distrl,digirl,yang2025zerogui}. Agent-based evaluators rely on the reasoning abilities of language models and procedural rubrics~\cite{agentasajudge,wadhwa2025evalagent}, but generally assume that the evidence to be judged has already been provided. GUI task evaluation presents a different challenge. The decisive evidence may not be contained in the given trajectory and can instead be distributed across screenshots, application states, files, configuration values, and command outputs~\cite{Wang2024c,Zheng2024a,Ye2026,Liu2024b}. IRA therefore treats evidence acquisition itself as part of evaluation. It first derives completion conditions, then determines which environment states and tools are relevant, and actively gathers evidence to verify each condition. This distinguishes IRA from prior agent-based judges that reason over fixed observations and grounds IRA's final judgment in evidence collected from the actual post-execution environment.

%% file: sections/method.tex
\section{Interactive Reward Agent}
\label{sec:Interactive Reward Agent}

\begin{figure*}[!t]
    \centering
    \includegraphics[width=\linewidth]{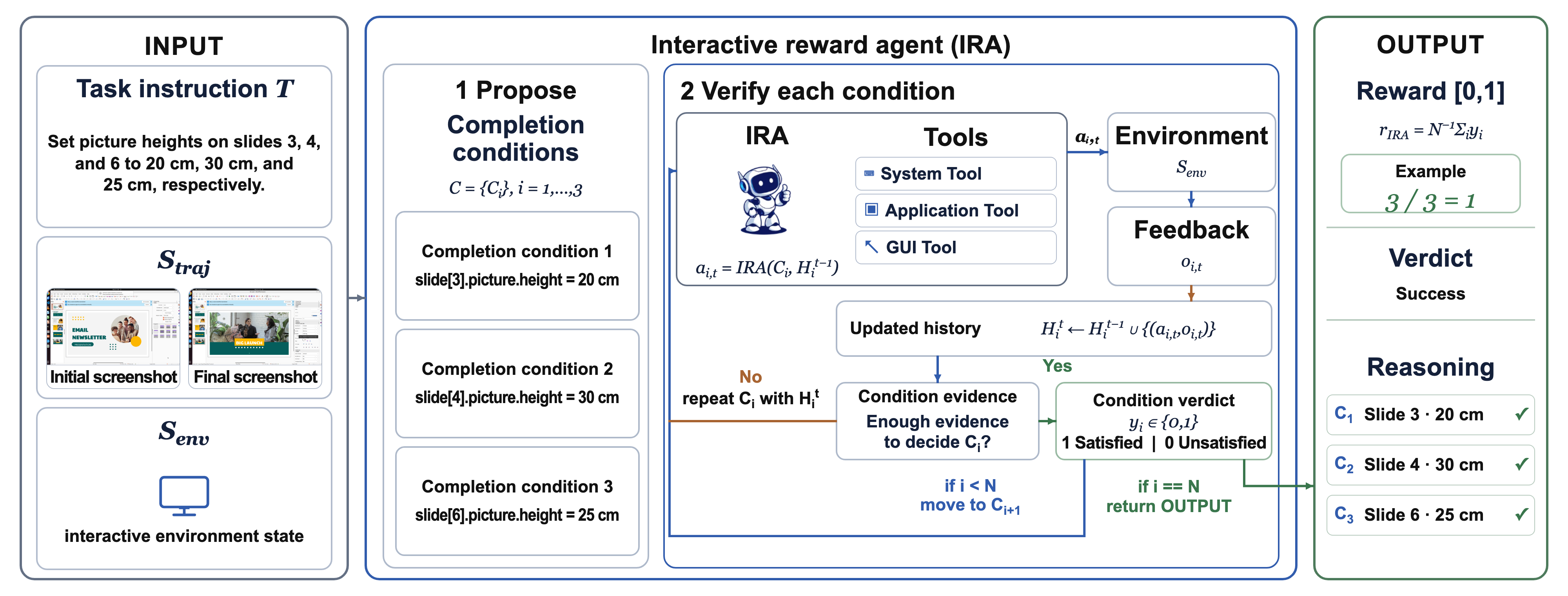}
    \caption{Overview of the Interactive Reward Agent (IRA). Given a task instruction \(T\), the initial and final screenshots \(I_{\mathrm{init}}\) and \(I_{\mathrm{final}}\), and an interactive environment state \(S_{\mathrm{env}}\), IRA first proposes a set of task-grounded completion conditions \(\{C_i\}_{i=1}^{N}\). For each condition, IRA iteratively selects tool actions based on the condition and interaction history, receives environment feedback, and updates the history until sufficient evidence is available to assign a binary verdict \(y_i\). The condition-level verdicts are then aggregated into the final reward, overall verdict, and reasoning.}
    \label{fig:pipeline}
\end{figure*}

\subsection{Background and Problem Setting}
Figure~\ref{fig:pipeline} shows the overall IRA pipeline. 
Given a task instruction $T$, screenshot evidence $S_{\mathrm{traj}}$, and the post-execution environment state $S_{\mathrm{env}}$, GUI task evaluation aims to predict a reward $r\in[0,1]$, where $r$ indicates the degree to which the task has been completed. We jointly denote the available evaluation evidence as $S=(S_{\mathrm{traj}},S_{\mathrm{env}})$, where $S_{\mathrm{traj}}=(I_{\mathrm{init}},I_{\mathrm{final}})$ contains the initial and final screenshots, and $S_{\mathrm{env}}$ denotes interactive environment states, such as files, configuration values, application states, command outputs, system settings, and structured interface information such as accessibility trees. Evaluation paradigms differ mainly in which part of $S$ they can access and how they verify task completion based on $S$.

Script-based evaluators assign a reward $r_{\mathrm{script}}$ using manually annotated task-specific verification functions $F_T$ in scripts, which can be formulated as
\begin{equation}
r_{\mathrm{script}} = F_T(S_{\mathrm{env}}),
\end{equation}
where $F_T$ contains predefined checking rules for task $T$.  This paradigm can provide accurate verification when task completion conditions are clear, but it requires separate scripts for different tasks.

VLM-based evaluators assign a reward $r_{\mathrm{vlm}}$ using a vision-language model $G_{\theta}$ based on visible observations, which can be formulated as
\begin{equation}
r_{\mathrm{vlm}} = G_{\theta}(T,S_{\mathrm{traj}}),
\end{equation}
where $G_{\theta}$ predicts the task completion reward conditioned on the instruction $T$ and the passive trajectory evidence $S_{\mathrm{traj}}$. This paradigm is more general across tasks, but its verification is mainly limited to information that is visually observable or implicitly inferable from $S_{\mathrm{traj}}$, making its reward prediction potentially inaccurate for tasks that require non-visual evidence or complex reasoning.

\subsection{Propose-then-Verify Framework}
IRA combines the scalability of VLM-based evaluators with the evidence-grounded reliability of script-based evaluators. Instead of directly predicting whether the task is complete, the agent first proposes a set of completion conditions from the task instruction $T$ and the initial and final screenshots $I_{\mathrm{init}}$ and $I_{\mathrm{final}}$:
\begin{equation}
C=P_{\theta}(T,I_{\mathrm{init}},I_{\mathrm{final}})=\{C_i\}_{i=1}^{N},
\end{equation}
where $P_{\theta}$ denotes the condition proposer.
Each condition $C_i$ describes one concrete requirement that should hold in the final environment state, such as the existence of an output file, the value of a configuration entry, the state of an application, or the visibility of a GUI element.

After proposing the conditions, the agent verifies each one through a ReAct-style reasoning-action-observation process. For a condition $C_i$, verification is not treated as a single-step query. Instead, the agent maintains a condition-specific interaction history $H_i^t$, where $t$ denotes the current verification step. At step $t$, the agent reasons over the condition and the accumulated history, decides what additional evidence is needed, and selects a tool action:
\begin{equation}
a_{i,t} = \mathrm{IRA}(C_i, H_i^{t-1}).
\end{equation}
The selected tool then queries the final environment state and returns an observation:
\begin{equation}
o_{i,t} = \mathrm{Tool}(a_{i,t}, S_{\mathrm{env}}).
\end{equation}
For read-only tools, the environment state remains unchanged, whereas GUI tools or executable commands may cause state transitions.
The observation is appended to the history:
\begin{equation}
H_i^{t}=H_i^{t-1}\oplus(a_{i,t},o_{i,t}),
\end{equation}
where $\oplus$ denotes ordered concatenation.

If the returned evidence is insufficient or ambiguous, the agent does not immediately mark the condition as satisfied. Instead, it searches for an alternative evidence source and issues another tool call. In this way, verification is an iterative process: the observation from one tool call determines the next reasoning step and the next tool action.

Let $K_i$ denote the number of verification steps performed for condition $C_i$. The final judgment for this condition is computed from the accumulated verification history:
\begin{equation}
y_i=\mathrm{IRA}(C_i,H_i^{K_i})\in\{0,1\}.
\end{equation}
We set $y_i=1$ only when the accumulated observations in $H_i^{K_i}$ provide explicit environment-state evidence that $C_i$ is satisfied; otherwise, we set $y_i=0$.

The final reward is computed by aggregating condition-level judgments across all conditions:
\begin{equation}
r_{\mathrm{ira}} = \frac{1}{N}\sum_{i=1}^{N} y_i .
\end{equation}
This scalar reward represents the fraction of satisfied conditions. For binary evaluation, rewards greater than 0.8 are classified as success.

\subsection{Evidence Acquisition and Tool Design}

IRA obtains explicit evidence through three complementary tool groups: system tools inspect files, configurations, command outputs, and structured interface state; application tools verify generated artifacts such as documents, spreadsheets, and presentations; and GUI tools expose otherwise inaccessible interface states through interactive navigation. During verification, IRA dynamically selects the tool that best matches each completion condition, incorporates the returned evidence into its interaction history, and updates its judgment. Supplementary Sec.~\ref{sec:appendix_tools} provides the core tool summary, complete specifications, and application-specific extensions.

%% file: sections/benchmark.tex
\section{GUI-RewardBench}
\label{sec:GUI-RewardBench}

\begin{figure*}[t]
\centering

\begin{minipage}[t]{0.62\linewidth}
\centering
\includegraphics[width=\linewidth]{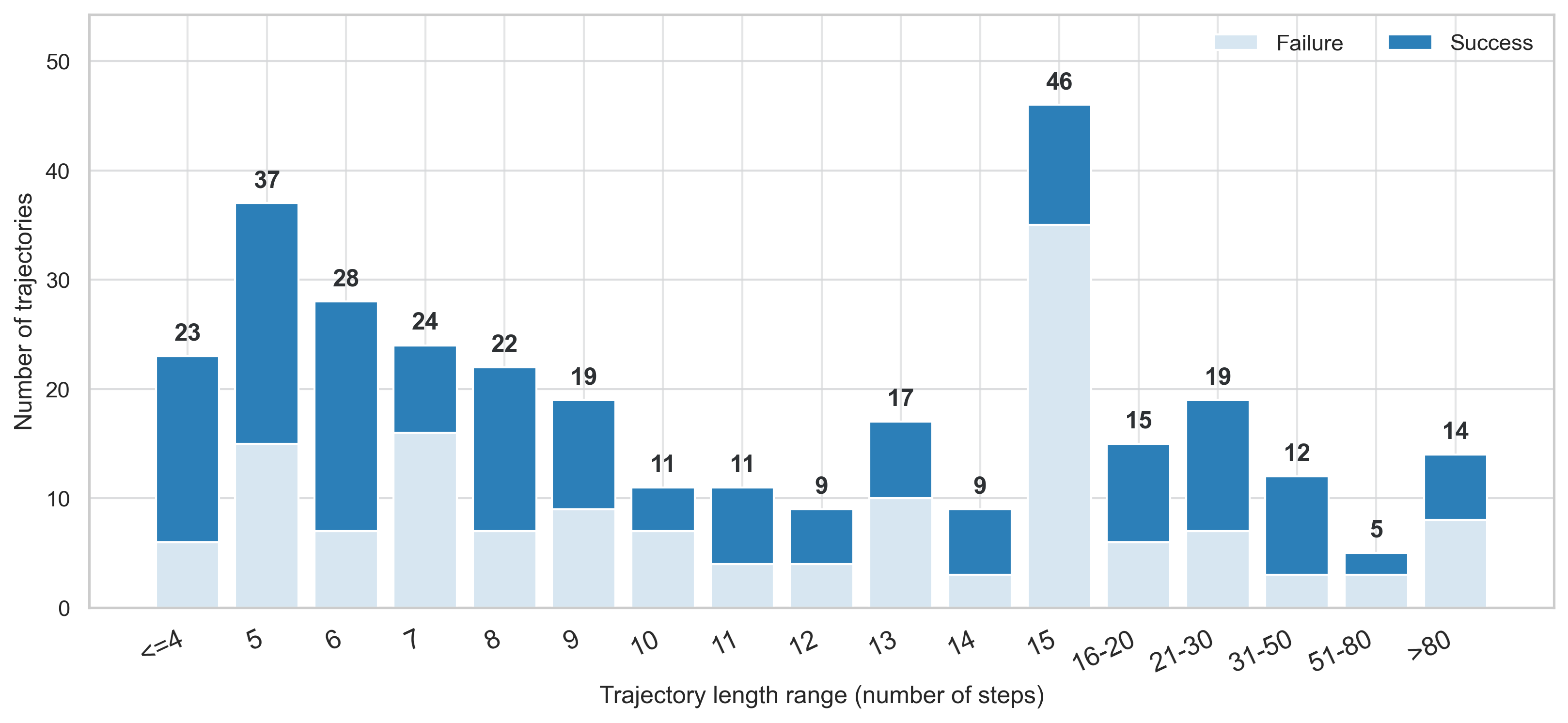}
\caption{Distribution of trajectory length and outcome (success or failure).}
\label{fig:rewardbench_length}
\end{minipage}
\hfill
\begin{minipage}[t]{0.33\linewidth}
\centering
\includegraphics[width=\linewidth]{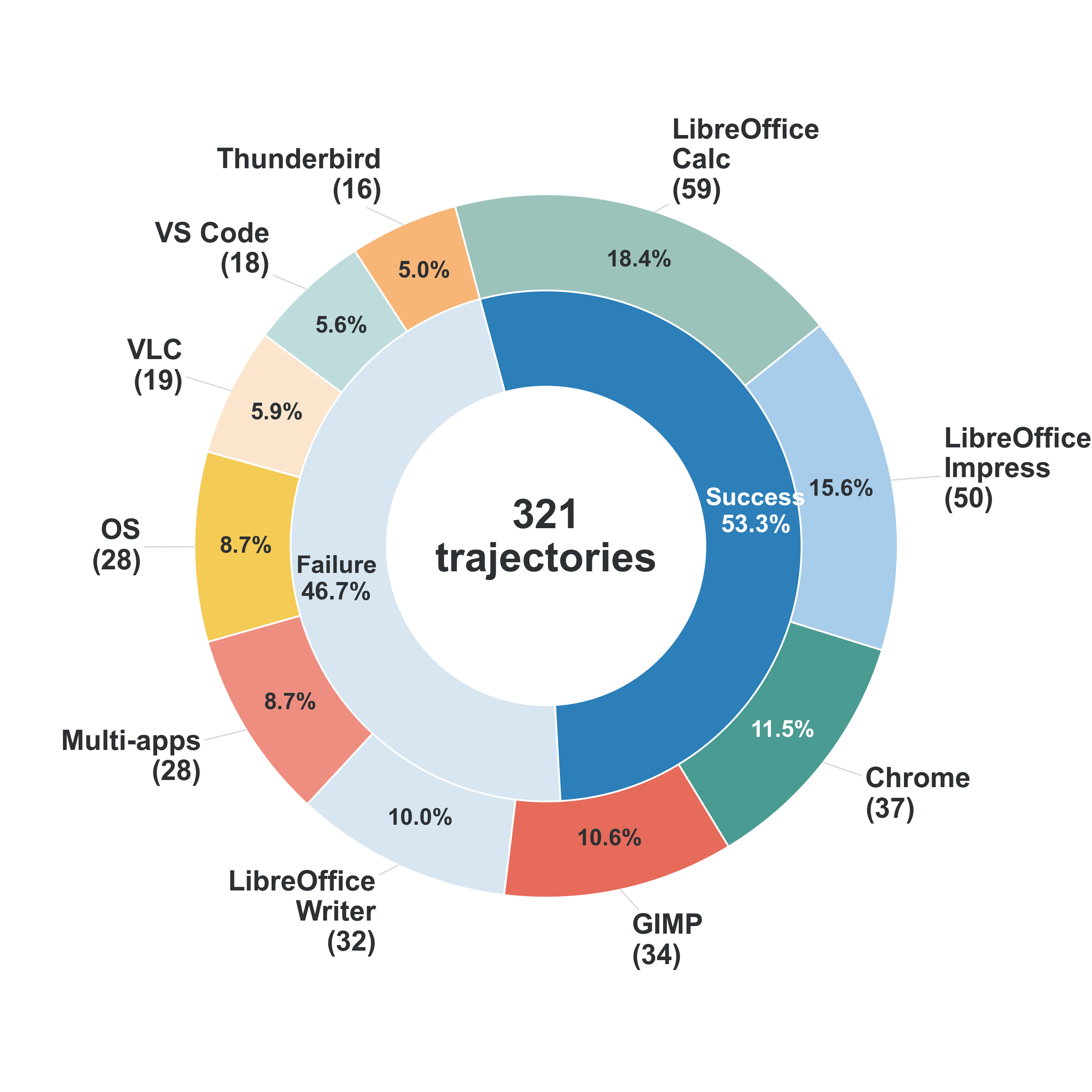}
\caption{Task-category distribution (outer ring) and success/failure distribution (inner ring).}
\label{fig:rewardbench_category}
\end{minipage}

\end{figure*}

GUI-RewardBench is designed to evaluate the accuracy of reward evaluators in judging GUI task completion from environment-state evidence. GUI-RewardBench emphasizes cases where completion evidence may be distributed across screenshots, files, application preferences, system settings, and generated artifacts.

\subsection{Benchmark Construction}
\label{sec:rewardbench_construction}

We construct GUI-RewardBench in Ubuntu desktop environments through trajectory replay. The task sources include OSWorld-derived tasks and manually created tasks that cover common desktop applications and multi-application workflows. The final benchmark contains 321 labeled task trajectories.

For each candidate task, we use UI-TARS-1.5~\cite{Qin2025} and EvoCUA-8B~\cite{xue2026evocua} to generate trajectories. 
We then replay the recorded action sequence across multiple virtual-machine instances and compare the resulting final states. A trajectory is included only when the replay produces a relatively stable final state. We initially sample 327 candidates, replay them three times, and remove 6 unstable ones, resulting in 321 stable trajectories. 

\subsection{Trajectory-Length and Outcome Distribution}
\label{sec:rewardbench_trajectory_distribution}

As shown in Fig.~\ref{fig:rewardbench_length}, GUI-RewardBench contains a relatively balanced mix of successful and failed trajectories, with diverse failure modes. Success trajectories satisfy the requested instruction, while failure trajectories include incorrect executions, partial completions, missing artifacts, wrong preference values, and visually plausible but state-inconsistent outcomes. Each benchmark trajectory is paired with the task-specific evaluation script for its underlying task.

\subsection{Benchmark Statistics}
\label{sec:rewardbench_statistics}

GUI-RewardBench consists of 321 stable evaluation trajectories across 10 Ubuntu desktop application categories. As shown in Fig.~\ref{fig:rewardbench_length}, the benchmark covers a broad range of trajectory lengths and workflow complexities. Trajectory lengths range from fewer than 5 steps to more than 80 steps, with 50 trajectories requiring over 20 steps. This distribution reflects both short interface-level operations and longer workflows involving file editing, export, and cross-application interaction.

The category distribution, shown in Fig.~\ref{fig:rewardbench_category}, covers office productivity, graphics editing, web browsing, communication, media playback, development, OS-level configuration, and multi-application workflows. Beyond the category diversity, GUI-RewardBench is characterized by three verification types. Artifact-verification tasks form the largest group, accounting for 192 tasks, and require inspecting generated documents, spreadsheets, presentations, images, PDFs, and other output files. Hidden-state tasks account for 89  tasks, requiring evidence beyond the visible screenshot, such as application preferences, configuration files, default handlers, or system settings. Visible-state tasks comprise the remaining 40 tasks and can typically be assessed directly from the interface.

This composition makes GUI-RewardBench suitable for evaluating desktop task verification beyond screenshot-level evidence. In particular, it stresses persistent state inspection, artifact correctness, and cross-application reasoning, which are essential for reliably judging real-world desktop workflows.

\subsection{Evaluation Protocol}
\label{sec:rewardbench_protocol}

All evaluation methods receive the same task instruction and evaluate the environment state produced by the same evaluation pipeline: environment initialization, trajectory replay, evaluator inspection, and task-specific script verification. Each method outputs a scalar reward, which is then converted to a binary prediction using the same threshold. Rewards greater than 0.8 are classified as success. Success is treated as the positive class when computing precision, recall, and F1. Although GUI-RewardBench trajectories are selected to be relatively replay-stable, evaluation is still conducted in a live desktop environment. UI drift, asynchronous application behavior, network variability, replay imperfections, and evaluator interactions may cause the replayed state to differ from the originally recorded state. Therefore, after each live evaluation run, we execute the task-specific evaluation script and use its output as the ground-truth label for computing TP, FP, TN, and FN. This protocol compares each evaluator against the actual environment state it inspected, while keeping the task instruction, recorded trajectory, replay procedure, and verification script fixed across methods. 

%% file: sections/experiments.tex
\section{Experiments}
\label{sec:experiments}

\begin{table*}[t]
  \centering
  \caption{Performance of VLM-based reward evaluators and the interactive reward agent on GUI-RewardBench.}
  \label{tab:overall_category_results}
  \small
  \resizebox{0.8\textwidth}{!}{
  \begin{tabular}{@{}llcccccccc@{}}
    \hline
    \textbf{Method} & \textbf{Mode} & \textbf{Acc.}$\uparrow$ & \textbf{Prec.}$\uparrow$ & \textbf{Rec.}$\uparrow$ & \textbf{F1}$\uparrow$ & \textbf{TP}$\uparrow$ & \textbf{FP}$\downarrow$ & \textbf{TN}$\uparrow$ & \textbf{FN}$\downarrow$ \\
   \hline
    \multicolumn{10}{@{}l}{\textit{VLM-based reward evaluators}} \\
    \addlinespace[1pt]
    WebRL~\cite{qi2025webrl} & \textsc{Passive} & 47.0\% & 100.0\% & 0.6\% & 1.2\% & 1 & 0 & 150 & 170 \\
    ZeroGUI~\cite{yang2025zerogui} & \textsc{Passive} & 67.6\% & 77.2\% & 55.6\% & 64.6\% & 95 & 28 & 122 & 76 \\
    DigiRL~\cite{digirl} & \textsc{Passive} & 78.5\% & 93.2\% & 64.3\% & 76.1\% & 110 & 8 & 142 & 61 \\
    DistRL~\cite{wang2025distrl} & \textsc{Passive} & 78.8\% & 92.6\% & 65.5\% & 76.7\% & 112 & 9 & 141 & 59 \\
    \hline
    \multicolumn{10}{@{}l}{\textit{Interactive reward agent (ours)}} \\
    \addlinespace[1pt]
    IRA with Qwen3.6-35B-A3B~\cite{qwen2026qwen36} & \textsc{Interactive} & 85.0\% & 88.7\% & \textbf{79.6\%} & 83.9\% & 125 & 16 & 148 & 32 \\
    IRA with GPT-5.4~\cite{openai2026gpt54} & \textsc{Interactive} & 85.4\% & 94.0\% & 76.2\% & 84.2\% & 125 & 8 & 149 & 39 \\
    IRA with GPT-5.5~\cite{openai2026gpt55} & \textsc{Interactive} & \textbf{86.9\%} & 93.7\% & 77.6\% & \textbf{84.9\%} & 118 & 8 & 161 & 34 \\
    \hline
  \end{tabular}}
\end{table*}

\subsection{Experimental Setup}
\label{sec:exp_setup}

\noindent \textbf{Comparison methods.}
We organize the comparison by evaluator formulation. Existing VLM-based reward evaluators, including WebRL~\cite{qi2025webrl}, ZeroGUI~\cite{yang2025zerogui}, DigiRL~\cite{digirl}, and DistRL~\cite{wang2025distrl}, are passive evaluators. They directly judge task completion from screenshots or visual trajectory evidence. 
IRA is an interactive evaluator and follows the propose-then-verify framework, allowing the evaluator to inspect environment state and use tools to verify task completion. We evaluate IRA using Qwen3.6-35B-A3B (hereafter Qwen3.6)~\cite{qwen2026qwen36}, GPT-5.4~\cite{openai2026gpt54}, and GPT-5.5~\cite{openai2026gpt55}. 

\noindent \textbf{Metrics and evaluation rules.}
We report accuracy, precision, recall, F1, TP, FP, TN, and FN. Following the evaluation protocol described above, we convert each scalar reward into a binary prediction, classifying rewards greater than 0.8 as success. Because interactive evaluators may alter the environment they inspect, the resulting positive/negative totals differ slightly across methods, even though the trajectory and verification script are held fixed.

\subsection{Quantitative Results}
\label{sec:main_quantitative_results}

Results are reported in 
Table~\ref{tab:overall_category_results}. The category-wise comparison is provided in Supplementary Sec.~\ref{sec:appendix_category_results}.

\noindent \textbf{IRA improves evaluation by collecting task-relevant evidence.}
All three IRA variants outperform all four passive VLM-based reward evaluators. The strongest passive evaluator, DistRL, achieves 78.8\% accuracy and 76.7\% F1, whereas every IRA variant achieves at least 85.0\% accuracy and 83.9\% F1. A VLM-based reward evaluator must infer task completion from fixed visual observations, which may not show whether a file was saved correctly, a setting was updated, or an output was produced in another application. IRA instead checks these states directly through tools. The results show that accurate GUI task evaluation depends not only on the VLM's reasoning ability but also on the evaluator's access to the evidence needed to verify task completion.

\noindent \textbf{IRA performs consistently across the three evaluated backbones.}
VLM-based reward evaluators rely on the backbone model to judge task completion directly from visual observations. Their performance is therefore limited by how well the backbone generalizes beyond its training domain. For example, WebRL~\cite{qi2025webrl} is designed for web-related tasks, which may contribute to its low recall on GUI-RewardBench. In contrast, IRA performs consistently across the three evaluated backbones. The open-source Qwen3.6 backbone achieves results close to the proprietary GPT-5.4 and GPT-5.5 backbones and obtains the highest recall. These results show that IRA reduces the dependence on the backbone's existing ability to generalize. By giving different models the same procedure for collecting and verifying evidence, IRA enables both open-source and proprietary backbones to serve as effective GUI task evaluators.

\subsection{Qualitative Results}
\label{sec:qualitative_results}

\begin{figure}[t]
    \centering
    \includegraphics[width=0.92\linewidth]{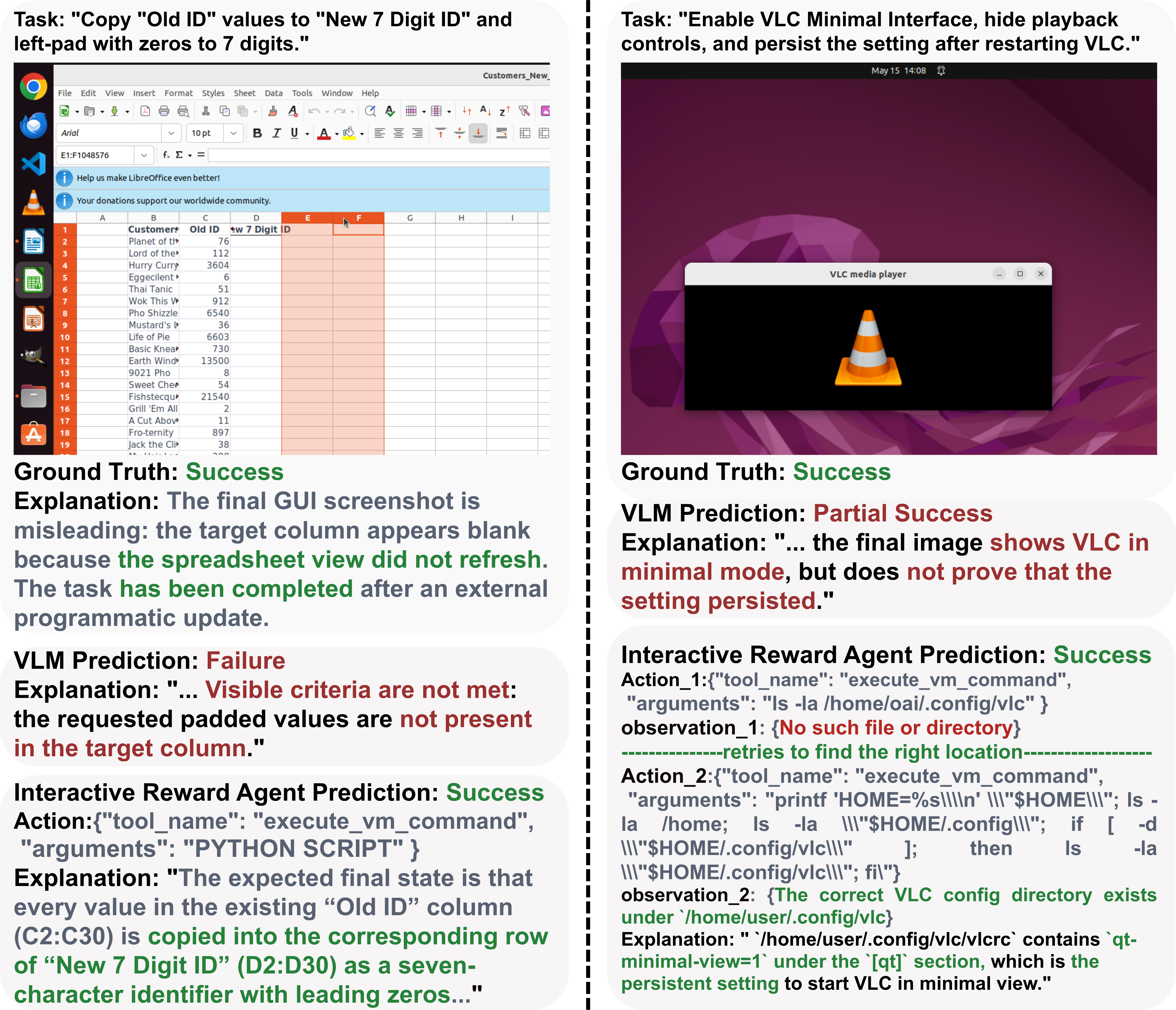}
    \caption{Qualitative comparisons between a VLM-based reward evaluator and the interactive reward agent (IRA) using the same GPT-5.5 backbone. In both cases, IRA produces the correct success verdict while the screenshot-only evaluator does not.}
    \label{fig:qualitative_ira}
\end{figure}

Figure~\ref{fig:qualitative_ira} compares a VLM-based reward evaluator and IRA using the same GPT-5.5 backbone. In the LibreOffice Calc example, the task is completed through a command-line update, but the spreadsheet view does not refresh. The VLM evaluator therefore predicts failure from the outdated screenshot, whereas IRA inspects the spreadsheet file and verifies the updated values. In the VLC example, the screenshot shows the current interface mode but cannot confirm whether the setting has been saved. IRA instead locates the configuration file and checks the stored value. These examples show that IRA can verify task states that are missing or unclear in screenshots. Additional examples and full trajectories are provided in Supplementary Secs.~\ref{sec:appendix_qualitative_results} and~\ref{sec:appendix_full_trajectories}. IRA's remaining errors mainly arise from misaligned completion conditions, including incorrect granularity, overly literal interpretations, and missed persistence requirements. Detailed cases are provided in Supplementary Sec.~\ref{sec:appendix_error_cases}.

\begin{table*}[t]
\centering
\caption{Accuracy and verification cost under matched backbones. IRA token counts are reported as median/mean.}
\label{tab:efficiency}
\scriptsize
\resizebox{0.88\textwidth}{!}{
\begin{tabular}{lcccccc}
\toprule
\textbf{Backbone}
& \textbf{VLM Acc.}
& \textbf{IRA Acc.}
& \textbf{$\Delta$ Acc. (pp)}
& \textbf{IRA Avg. Steps}
& \textbf{VLM Tokens}
& \textbf{IRA Tokens (Med./Mean)} \\
\midrule
GPT-5.5 & 78.5\% & \textbf{86.9\%} & $+8.4$ & 3.34 & 3.05K & \textbf{14.9K / 59.5K} \\
GPT-5.4 & 76.3\% & 85.4\% & \textbf{$+9.1$} & \textbf{3.27} & 3.11K & 23.8K / 120.0K \\
Qwen3.6-35B-A3B & 78.8\% & 85.0\% & $+6.2$ & 8.20 & 2.98K & 25.9K / $\sim$139.1K \\
\bottomrule
\end{tabular}}
\end{table*}

\subsection{Ablation Study}
\label{sec:ablation_study}
\begin{figure}
    \centering
    \includegraphics[width=0.8\linewidth]{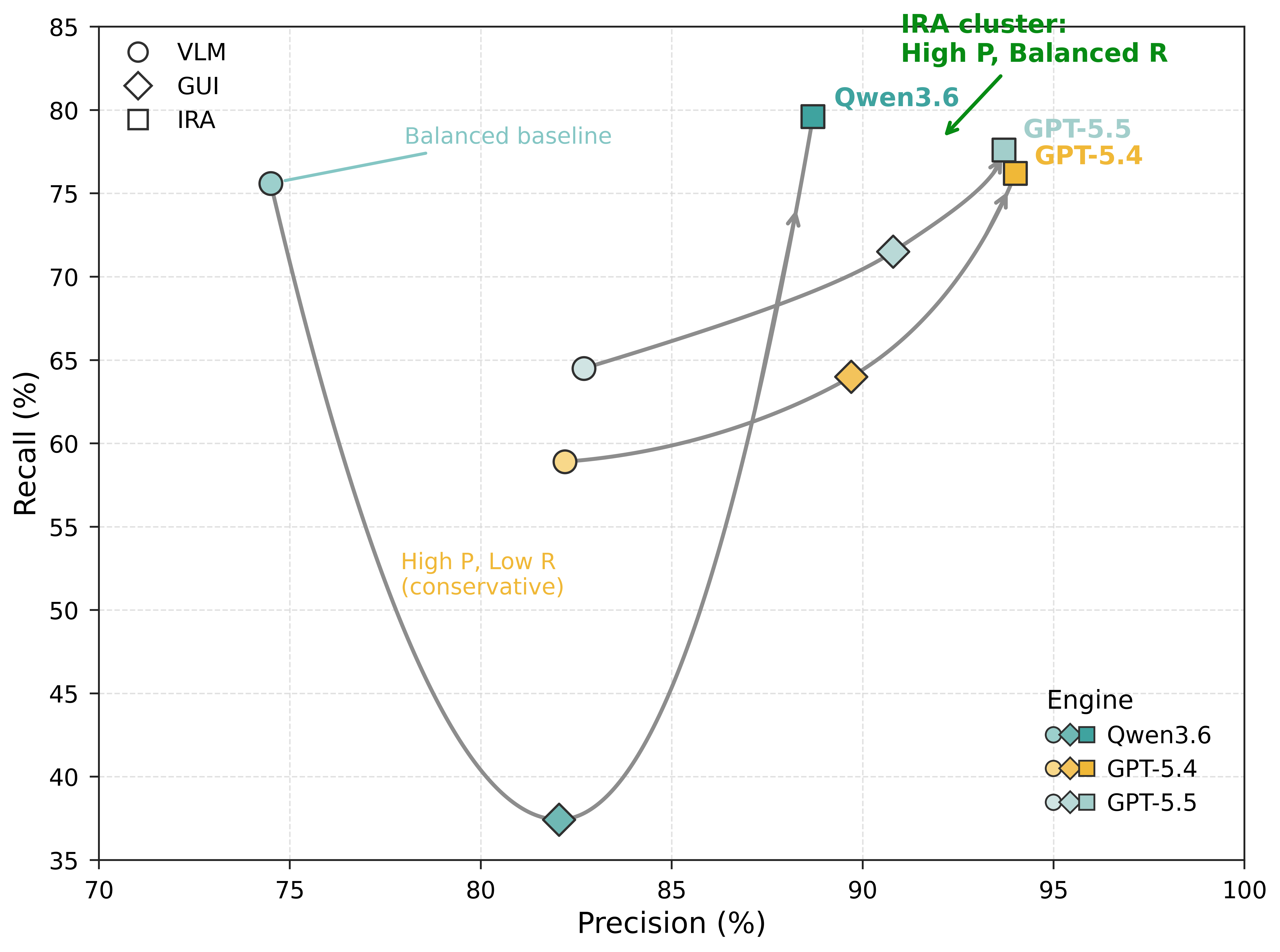}
    \caption{Precision--recall trajectories under matched backbones.
    Each curve connects the VLM-only, GUI-only, and IRA settings for the same backbone. The VLM-only setting judges task completion directly from the initial and final screenshots; the GUI-only setting adds environment interaction; and IRA uses the full propose-then-verify framework with GUI, system, and application tools. Circles, diamonds, and squares denote VLM-only, GUI-only, and IRA, respectively.}
    \label{fig:formulation_ablation}
    \label{fig:placeholder}
\end{figure}

We compare three evaluator formulations under the same backbone. In the VLM-only setting, the evaluator receives the task instruction and the initial and final screenshots, and directly predicts whether the task is completed. The GUI-only setting additionally allows the evaluator to interact with the environment to make a judgment. The IRA setting uses the full propose-then-verify framework. Fig.~\ref{fig:formulation_ablation} reports the resulting precision and recall under matched backbones.

\noindent \textbf{In these experiments, GUI interaction helps only when the backbone uses it reliably.}
For GPT-5.4 and GPT-5.5, adding GUI exploration improves both precision and recall, showing that active inspection can reveal evidence missing from the input screenshots. Qwen3.6 shows a different pattern. Its precision improves, but its recall drops sharply, indicating a more conservative evaluation mode. GUI-only verification requires the backbone to plan interactions, ground interface elements, recover from errors, and interpret the resulting state. When these steps are unreliable, failure to find evidence may be mistaken for evidence of task failure, making GUI interaction less effective than direct visual judgment.

\noindent \textbf{IRA turns interaction into structured verification.}
The full IRA setting improves the precision--recall trade-off across all three backbones and reverses the degradation observed with GUI-only Qwen3.6. This contrast shows that interaction access alone is not sufficient. IRA first defines the completion conditions and then selects system, application, or GUI tools to verify each condition with explicit evidence. By reducing reliance on open-ended and error-prone GUI exploration, IRA enables the open-source Qwen3.6 backbone to perform strongly, highlighting the importance of the verification framework and tool interface.

\subsection{Verification Cost and Tool Usage}
\label{sec:tool_usage_efficiency}

Table~\ref{tab:efficiency} shows that IRA improves accuracy by $6.2$--$9.1$ percentage points. The two GPT backbones obtain these gains in about $3.3$ verification steps on average, while Qwen3.6 uses more steps and a heavier-tailed token budget. This additional cost provides environment evidence that static visual reasoning cannot recover, including file contents, saved configurations, and application state. The GPT backbones also use fewer tool calls per task (2.34 and 2.53) than Qwen3.6 (7.31), favoring a small number of command-oriented checks over broader GUI exploration. Complete tool-level distributions are reported in Supplementary Sec.~\ref{sec:appendix_tool_usage}.

\subsection{Automated Task Generation and Human--IRA Agreement}
\label{sec:autotaskgen}

To extend evaluation beyond GUI-RewardBench, we group validated tasks by their initial environment configuration and use an LLM to generate feasible instruction variants, producing 855 tasks across nine application categories. We randomly sample 100 of these tasks, execute them with a Qwen3.6-based GUI agent, and compare GPT-5.5-based IRA verdicts against human annotations. IRA agrees with human labels on 94.0\% of the trajectories, yielding Cohen's $\kappa=0.84$. See Supplementary Secs.~\ref{sec:appendix_autotaskgen} and~\ref{sec:appendix_human_agreement} for more details of the generation pipeline and human-IRA agreement experiments.

\subsection{Reinforcement Learning with IRA Rewards}

\begin{minipage}[t]{\linewidth}
\centering
\captionof{table}{RL training settings and resulting OSWorld success rates.}
\label{tab:rl_settings}
\small
\resizebox{\linewidth}{!}{
\begin{tabular}{lllc}
\toprule
\textbf{Setting} & \textbf{Task Source} & \textbf{Evaluator} & \textbf{Success (\%)} \\
\midrule
A & OSWorld tasks & Script & 34.9 \\
B & OSWorld tasks & IRA & 34.0 \\
C & Generated tasks & IRA & 33.5 \\
\bottomrule
\end{tabular}}
\end{minipage}

We test whether IRA can replace task-specific reward scripts and extend RL training to automatically generated tasks.
Setting A trains on OSWorld tasks using ground-truth script rewards. Setting B uses the same tasks but replaces the scripts with IRA. Setting C trains on automatically generated tasks using IRA, where no task-specific evaluation script is available. 
We adopt the RL training method of DART~\cite{li2025efficient} and use Qwen3.6-35B-A3B as the backbone for IRA. 
Replacing script rewards with IRA on the same OSWorld tasks yields 34.0\% success, comparable to 34.9\%. More notably, training on generated out-of-distribution (OOD) tasks with IRA achieves 33.5\%, only 1.4 percentage points below the script-based setting. These results show that IRA converts generated tasks without evaluation scripts into effective, transferable RL supervision, enabling scalable training beyond curated benchmark tasks.

%% file: sections/conclusion.tex
\section{Discussion and Conclusion}

Our results indicate that reliable GUI reward modeling is primarily an evidence-verification problem rather than merely a visual classification problem. IRA separates task interpretation from evidence acquisition by decomposing GUI instructions into explicit completion conditions and verifying them through system, application, and GUI tools. This allows IRA to inspect evidence that passive evaluators cannot observe. Its consistent gains across backbones, together with the ablation and efficiency results, show the benefit of selecting decisive evidence. On the 321-trajectory GUI-RewardBench, IRA reaches 86.9\% accuracy with GPT-5.5 and strong human agreement on automatically generated tasks ($\kappa=0.84$). As an RL reward evaluator, it achieves a 34.0\% OSWorld success rate, close to the 34.9\% obtained with ground-truth script rewards. These results support environment-state verification as a practical approach to scalable GUI-agent evaluation and training.

%% file: supplementary_sections/appendix.tex
\setcounter{secnumdepth}{2}
\appendix
\section{Appendix}
\suppressfloats[t]
\label{sec:appendix}

This appendix provides implementation details, complete tool specifications, system prompts, hyperparameters, configurations, and baseline implementation details.

\subsection{Complete Tool Set}
\label{sec:appendix_tools}
The interactive reward agent (IRA) can support a much larger tool set spanning system-level inspection, application-specific verification, and GUI interaction capabilities. However, the complete tool set is not loaded for every evaluation episode.

During the experiments, only a subset of tools was activated depending on the target application. System tools and GUI interaction tools were always available, while application-specific tools were loaded on demand. In particular, \texttt{check\_word\_file}, \texttt{check\_excel\_file}, \texttt{check\_ppt\_file}, and \texttt{get\_ppt\_xml} were only enabled for the corresponding LibreOffice Writer, Calc, and Impress tasks. The full tool set was available only in multi-application settings where multiple software environments could be involved simultaneously.

The complete tool set includes several additional application-specific tools that were designed to provide structured access to application states and reduce the reliance on visual reasoning. These tools were introduced as optional capabilities to offload potentially expensive VLM-based inspection and verification steps. In practice, however, our experiments showed that IRA was often able to complete evaluations effectively using a much smaller core tool set. Consequently, many application-specific tools were invoked infrequently or not at all in the reported benchmarks.

This observation reflects a broader design principle of IRA: maintaining a minimal and general-purpose tool interface whenever possible, while retaining specialized tools as optional extensions for applications that may benefit from additional structured access.

\begin{table}[t]
\caption{Core interactive reward agent (IRA) tool set used in the main experiments.}
\label{tab:app_core_tools}
\centering
\scriptsize
\begin{tabular}{lp{5.5cm}}
\toprule
\textbf{Tool} & \textbf{Purpose} \\
\midrule
\multicolumn{2}{l}{\textit{System Tools (Generic)}} \\
execute\_vm\_command & Execute shell commands to check system state \\
get\_vm\_file & Retrieve file contents from the virtual machine \\
get\_terminal\_output & Access command history and outputs \\
get\_host\_file\_content & Read host-side file contents \\
get\_accessibility\_tree & Inspect UI structure, labels, roles, and values \\
\midrule
\multicolumn{2}{l}{\textit{Application Tools}} \\
check\_word\_file & LibreOffice Writer document verification \\
check\_excel\_file & LibreOffice Calc spreadsheet verification \\
check\_ppt\_file & LibreOffice Impress presentation verification \\
get\_ppt\_xml & PPT XML structure extraction \\
\midrule
GUI Tools & Interactive GUI navigation, including clicking, typing, and scrolling \\
\bottomrule
\end{tabular}
\end{table}

\begin{itemize}

\item \textbf{System Tools }
    \begin{itemize}
        \item \texttt{execute\_vm\_command}: Execute shell commands to inspect system state, check processes, verify system settings, or query command outputs.
        
        \textit{Parameters:} \texttt{command} (string)

        \item \texttt{get\_vm\_file}: Read file contents from the virtual machine filesystem. Useful for inspecting configuration files, generated outputs, logs, and text artifacts.
        
        \textit{Parameters:} \texttt{file\_path} (string)

        \item \texttt{get\_terminal\_output}: Retrieve previous terminal command outputs when task success is reflected in command-line feedback.

        \item \texttt{get\_directory\_listing}: List files and directories at a specified path. Verify file existence and check directory structure.
        
        \textit{Parameters:} \texttt{directory\_path} (string)
    \end{itemize}

\item \textbf{Chrome Tools}
    \begin{itemize}
        \item \texttt{get\_bookmarks}: Retrieve browser bookmark list with titles and URLs.
        \item \texttt{get\_browser\_history}: Access browsing history entries with URLs, titles, and timestamps.
        \item \texttt{get\_active\_tab\_info}: Get current active tab's URL, title, and status.
        \item \texttt{get\_cookie\_data}: Inspect browser cookies for a specified domain.
        
        \textit{Parameters:} \texttt{domain} (string)
        
        \item \texttt{get\_default\_search\_engine}: Query the currently configured default search engine.
    \end{itemize}

\item \textbf{Thunderbird Tools}
    \begin{itemize}
        \item \texttt{get\_thunderbird\_prefs}: Read Thunderbird preferences file (\texttt{prefs.js}) to verify configuration values.
        \item \texttt{get\_thunderbird\_timezone}: Check the configured timezone setting in Thunderbird.
        \item \texttt{get\_thunderbird\_accounts}: List all configured email accounts with associated settings.
    \end{itemize}

\item \textbf{VLC Tools}
    \begin{itemize}
        \item \texttt{get\_vlc\_config}: Read VLC configuration file to inspect settings such as recording path, subtitle preferences, and audio defaults.
        \item \texttt{get\_default\_video\_player}: Query the system's default video player application.
        \item \texttt{get\_vlc\_playing\_info}: Retrieve current playback state, including file path, play/pause status, and position.
    \end{itemize}

\item \textbf{VSCode Tools}
    \begin{itemize}
        \item \texttt{get\_vscode\_settings}: Read VSCode \texttt{settings.json} to verify editor configurations, theme, font size, and keybindings.
        \item \texttt{get\_vscode\_extensions}: List all installed VSCode extensions with names, versions, and status.
    \end{itemize}

\item \textbf{LibreOffice Tools}
    \begin{itemize}
        \item \texttt{check\_word\_file}: Verify LibreOffice Writer documents (\texttt{.odt}, \texttt{.docx}) by extracting text content, checking formatting, and inspecting document structure.
        
        \textit{Parameters:} \texttt{file\_path} (string)

        \item \texttt{check\_excel\_file}: Verify LibreOffice Calc spreadsheets (\texttt{.ods}, \texttt{.xlsx}) by reading cell values, formulas, and sheet structure.
        
        \textit{Parameters:} \texttt{file\_path} (string)

        \item \texttt{check\_ppt\_file}: Verify LibreOffice Impress presentations (\texttt{.odp}, \texttt{.pptx}) by extracting slide content, text, and layout structure.
        
        \textit{Parameters:} \texttt{file\_path} (string)
    \end{itemize}

\item \textbf{GIMP Tools}
    \begin{itemize}
        \item \texttt{get\_gimp\_config\_file}: Read GIMP configuration to verify tool settings, brush defaults, and preferences.
    \end{itemize}

\item \textbf{Interaction Tools}
    \begin{itemize}
        \item \texttt{observe\_current\_state}: Capture a screenshot of the current GUI state and optionally query a vision language model about specific visual elements.
        
        \textit{Parameters:} \texttt{query} (string, optional)

        \item \texttt{get\_accessibility\_tree}: Retrieve a structured accessibility tree showing UI elements, roles, labels, values, and states.

        \item \texttt{GUI Tools}: Perform interactive GUI actions including mouse clicks, keyboard input, scrolling, and navigation.
        
        \textit{Parameters:}
        \begin{itemize}
            \item \texttt{action} (string)
            \item \texttt{coordinate} ([x,y])
            \item \texttt{text} (string)
            \item \texttt{keys} (list)
            \item \texttt{pixels} (int)
            \item \texttt{time} (float)
        \end{itemize}
    \end{itemize}

\item \textbf{Final Output Tool}
    \begin{itemize}
        \item \texttt{final\_answer}: Output the final evaluation result with reward score, verdict, and detailed reasoning. This tool must be called exactly once at the end of evaluation.
        
        \textit{Parameters:}
        \begin{itemize}
            \item \texttt{reward} (float, 0--1)
            \item \texttt{verdict} (Success / Partial Success / Failure)
            \item \texttt{reasoning} (string)
        \end{itemize}
    \end{itemize}

\end{itemize}

\subsection{System Prompt Structure} The IRA system prompt enforces a reasoning-action-observation loop and contains five key components:

\noindent\textbf{Workflow Structure.} The agent operates in a ReAct-style loop:

\begin{itemize}
\item \textbf{Thought} — The agent outputs reasoning in natural language: what condition to verify, what evidence is needed, which tool to use
\item \textbf{Action} — The agent calls exactly one tool via function calling API
\item \textbf{Observation} — The system executes the tool and returns the result
\end{itemize}

This pattern continues until all conditions are verified or the agent outputs a final answer.

\noindent\textbf{Evaluation Protocol.} The prompt enforces a five-step evaluation process:

\begin{itemize}
\item \textbf{Initial Analysis} — Understand task requirements, compare initial and final screenshots
\item \textbf{Verification Strategy} — Choose the strongest verification path for each condition
\item \textbf{Define Completion Conditions} — Decompose task into atomic, necessary completion conditions
\item \textbf{Verify Each Condition} — Use tools to gather evidence, mark each condition as satisfied/unsatisfied
\item \textbf{Calibrate Verdict} — Aggregate condition results into final reward
\end{itemize}

\noindent\textbf{Core Principles.} Three principles guide evaluation:

\begin{itemize}
\item \textbf{Final-State Only} — Judge based solely on final environment state; intermediate progress is ignored
\item \textbf{Task-Grounded Conditions} — Every condition must be strictly necessary for task completion; do not introduce requirements beyond task instruction
\item \textbf{Conservative Evaluation} — Ambiguous or unverifiable conditions are treated as NOT satisfied
\end{itemize}

\noindent\textbf{Evidence Selection Policy.} The prompt specifies a strict evidence hierarchy:

\begin{itemize}
\item For visible states → use attached screenshots
\item For application states → use specialized tools (e.g., \texttt{get\_bookmarks}, \texttt{get\_vscode\_settings})
\item For hidden states → use CLI/file tools (e.g., \texttt{execute\_vm\_command}, \texttt{get\_vm\_file})
\item For GUI inspection → use \texttt{computer} tool as a last resort
\end{itemize}

This hierarchy encourages the agent to use more reliable evidence sources first.

\noindent\textbf{Strict Prohibitions.} The prompt explicitly forbids common failure modes:

\begin{itemize}
\item Do NOT repeat tool calls after receiving results
\item Do NOT speculate beyond observable evidence
\item Do NOT skip writing Thought before Action
\item Do NOT make multiple tool calls per turn
\item Do NOT assume success without explicit verification
\item Do NOT add conditions beyond task requirements
\end{itemize}

\subsection{Hyperparameters and Configuration}
\label{sec:appendix_hyperparameters}

\noindent\textbf{Model Configuration.} We report details of model configurations in Table \ref{tab:model_configs}. 

\begin{table}[h]
\centering
\caption{Model configurations for IRA experiments.}
\label{tab:model_configs}
\small
\begin{tabular}{lp{5cm}}
\toprule
\textbf{Parameter} & \textbf{Value / Description} \\
\midrule
\textbf{Backbone Models} & GPT-5.5, GPT-5.4, Qwen3.6-35B-A3B\\
Temperature & 0.0  \\
Tool call format & OpenAI function calling API \\
Vision input & Screenshots encoded as base64 images \\
Screenshot retention & Max 5 most recent screenshots in context \\
Max evaluation steps & 30 \\
\midrule
\textbf{Retry Policy} & \\
Max retries & 3 (for API errors) \\
Backoff & Exponential (1s, 2s, 4s) \\
\bottomrule
\end{tabular}
\end{table}

\subsection{Baseline Implementation Details}
\label{sec:appendix_baselines}

\noindent\textbf{GUI-agent-as-reward evaluator Baselines}

\textbf{GPT-5.5 GUI-agent-as-reward evaluator:} Use GPT-5.5 with GUI tool access. Prompted as a verification agent that can interact with the GUI to check task completion. No structured condition-centric protocol.

\textbf{GPT-5.4 GUI-agent-as-reward evaluator:} Same as GPT-5.5 but with GPT-5.4 backbone.

\begin{minipage}[t]{\linewidth}
\centering
\captionof{table}{IRA tool usage across backbones. Percentages cover tool invocations; the final row reports the average number per task.}
\label{tab:tool_usage_pct}
\small
\resizebox{\linewidth}{!}{
\begin{tabular}{lrrr}
\toprule
\textbf{Tool / Category} & \textbf{GPT-5.5} & \textbf{GPT-5.4} & \textbf{Qwen3.6} \\
\midrule
execute\_vm\_command & 83.4\% & 65.6\% & 45.9\% \\
GUI tools & 6.4\% & 10.3\% & 50.5\% \\
get\_accessibility\_tree & 7.3\% & 16.3\% & 2.0\% \\
get\_terminal\_output & 0.0\% & 1.6\% & 0.6\% \\
get\_vm\_file & 1.1\% & 3.4\% & 0.3\% \\
Other tools & 1.9\% & 2.7\% & 0.7\% \\
\midrule
\textbf{Avg. tools per task} & \textbf{2.34} & \textbf{2.53} & \textbf{7.31} \\
\bottomrule
\end{tabular}}
\end{minipage}

\input{supplementary_sections/additional_experimental_details}

\onecolumn
\noindent

\subsection{Qualitative trajectory visualization}
\label{sec:appendix_full_trajectories}

We visualize representative IRA evaluation trajectories to illustrate how the agent reaches its final judgment. Each trajectory records the initial and final screenshots available to the visual reward model, followed by the IRA's step-by-step thoughts, tool calls, GUI interactions, observations, and final verdict. These examples highlight cases where the final screenshot alone is insufficient and the correct judgment depends on hidden document contents, spreadsheet cells, browser preferences, application configuration files, or cross-application evidence. 

\input{supplementary_sections/reward_agent_trajectory_appendix}

%% file: supplementary_sections/additional_experimental_details.tex
\subsection{Category-wise Evaluation Results}
\label{sec:appendix_category_results}

\begin{figure}[t]
    \centering
    \includegraphics[width=\linewidth]{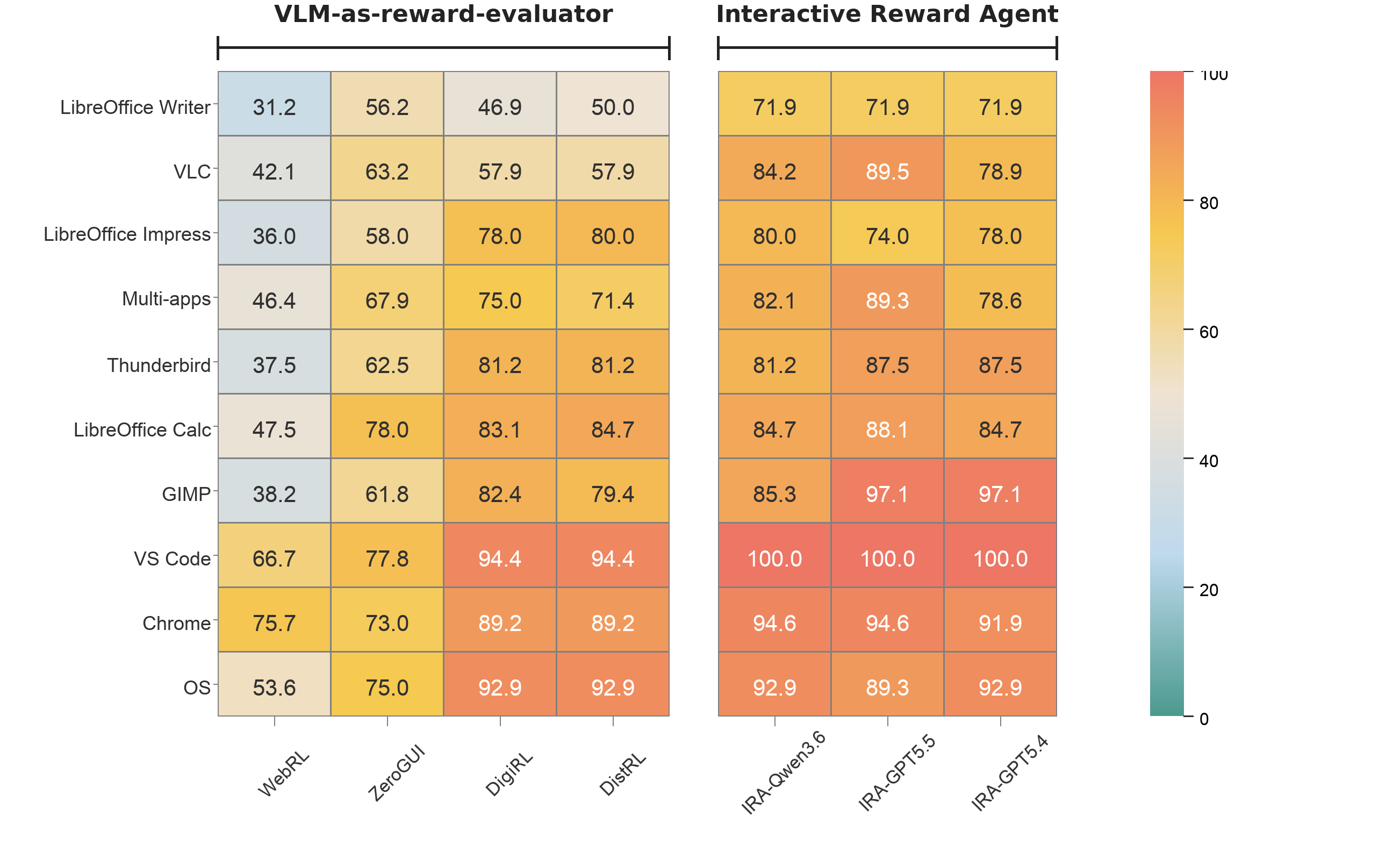}
    \caption{Per-category accuracy comparison between VLM-based reward evaluators and IRA. Each cell reports accuracy (\%) for one method on one application category, with warmer colors indicating higher accuracy.}
    \label{fig:category_heat}
\end{figure}

Figure~\ref{fig:category_heat} reports the per-category accuracy of the passive VLM-based reward evaluators and IRA variants. The category breakdown complements the overall results in the main paper and shows that IRA's improvement is distributed across diverse applications rather than being driven by a single category.

\subsection{Additional Qualitative Results}
\label{sec:appendix_qualitative_results}

The two qualitative comparisons summarized in the main paper use the same GPT-5.5 backbone for the VLM-based reward evaluator and IRA. In the LibreOffice Calc example, the task has been completed through a command-line update, but the visible spreadsheet view did not refresh. The screenshot-only evaluator therefore produces a false negative, whereas IRA queries the underlying file and verifies that the values were copied and zero-padded to seven digits. In the VLC example, the final screenshot shows the current interface mode but cannot establish whether the setting persists. IRA verifies the stored configuration value and searches for the correct path after its first command returns no matching file. These cases show how IRA converts incomplete visual observations into explicit evidence from files, configurations, and application states.

\subsection{Automated Task Generation Details}
\label{sec:appendix_autotaskgen}

To extend evaluation beyond the 321 trajectories in GUI-RewardBench, we develop an automated task generation pipeline to synthesize diverse GUI task instructions. The pipeline takes existing validated tasks and generates semantically similar variants through LLM-based instruction synthesis in two stages:

\begin{enumerate}
    \item \textbf{Configuration Extraction and Grouping.} Given a set of validated tasks, the pipeline extracts the \texttt{config} field from each task, which specifies the initial environment setup, including file downloads, application states, and system configurations. Tasks with identical configurations are grouped because they share the same starting environment state and can support similar instruction variants.
    \item \textbf{Instruction Generation.} For each configuration group, a language model generates new instructions that are feasible under the given configuration, require interactions similar to the reference tasks, test different aspects of the same application features, or maintain comparable complexity. The prompt includes the configuration specification, reference instructions, and application-specific constraints, and the model returns a JSON array of instruction candidates.
\end{enumerate}

The pipeline processes the GUI-RewardBench task configurations and generates 855 instructions across nine application categories. Table~\ref{tab:autogen_task_stats} summarizes their distribution. The unusually high number of Thunderbird tasks per configuration comes from one rich mail-profile initialization containing the required inbox messages, folders, attachments, account settings, and writable preference, filter, and contact data stores. These tasks vary mainly in their objectives rather than their initial environment.

\begin{table}[t]
\centering
\caption{Distribution of automatically generated tasks by category. The average is computed over tasks with identical initialization \texttt{config} fields.}
\label{tab:autogen_task_stats}
\small
\resizebox{\columnwidth}{!}{
\begin{tabular}{lcc}
\toprule
\textbf{Task Category} & \textbf{Generated Tasks} & \textbf{Avg. tasks/configuration} \\
\midrule
LibreOffice Calc & 227 & 5.4 \\
LibreOffice Writer & 105 & 4.8 \\
LibreOffice Impress & 204 & 4.6 \\
Chrome & 97 & 3.9 \\
VSCode & 24 & 4.0 \\
Thunderbird & 54 & 54.0 \\
VLC & 19 & 3.2 \\
GIMP & 33 & 2.5 \\
OS & 92 & 5.8 \\
\midrule
\textbf{Total} & \textbf{855} & \textbf{4.9} \\
\bottomrule
\end{tabular}}
\end{table}

\subsection{Human--IRA Agreement Details}
\label{sec:appendix_human_agreement}

We randomly sample 100 tasks from the 855 generated tasks, ensuring coverage across major application categories. For each task, a Qwen3.6-based GUI agent executes the instruction, IRA with GPT-5.5 evaluates the resulting trajectory, and human reviewers independently annotate task success. Human annotations serve as reference labels and IRA's binary verdicts as model labels.

The observed agreement rate is
\begin{equation}
p_o = \frac{n_{SS} + n_{FF}}{N},
\end{equation}
where $n_{SS}$ denotes tasks labeled successful by both human reviewers and IRA, $n_{FF}$ denotes tasks labeled as failures by both, and $N$ is the total number of evaluated tasks. Cohen's kappa adjusts this agreement for chance:
\begin{equation}
\kappa = \frac{p_o - p_e}{1 - p_e},
\end{equation}
where, for binary success/failure judgments,
\begin{equation}
p_e =
\frac{n_{S*}}{N}\frac{n_{*S}}{N}
+
\frac{n_{F*}}{N}\frac{n_{*F}}{N}.
\end{equation}
Here, $n_{S*}$ and $n_{F*}$ are the numbers of human success and failure labels, and $n_{*S}$ and $n_{*F}$ are the corresponding IRA verdict totals. The coefficient ranges from $-1$ (perfect disagreement) to 1 (perfect agreement), with 0 indicating chance-level agreement.

\begin{table}[t]
\centering
\caption{Contingency table between human reference labels and IRA verdicts.}
\label{tab:human_ira_contingency}
\small
\resizebox{\columnwidth}{!}{
\begin{tabular}{lccc}
\toprule
 & \textbf{IRA Success} & \textbf{IRA Failure} & \textbf{Total} \\
\midrule
\textbf{Human Success} & 23 & 6 & 29 \\
\textbf{Human Failure} & 0 & 71 & 71 \\
\midrule
\textbf{Total} & 23 & 77 & 100 \\
\bottomrule
\end{tabular}}
\end{table}

Table~\ref{tab:human_ira_contingency} gives an observed agreement rate of 94.0\% and $\kappa=0.84$, conventionally interpreted as almost perfect agreement. Table~\ref{tab:human_ira_kappa} reports the category-wise results. Cohen's $\kappa$ is N/A for Chrome because all sampled Chrome tasks are labeled as failures by both human reviewers and IRA; in this single-class case, $p_e=1$ and the denominator $1-p_e$ is zero.

\begin{table}[t]
\centering
\caption{Human--IRA agreement on generated tasks. ``LibreOffice'' combines Calc, Impress, and Writer.}
\label{tab:human_ira_kappa}
\small
\begin{tabular}{lccc}
\toprule
\textbf{Category} & \textbf{Sample Size} & \textbf{Agreement \%} & \textbf{$\kappa$} \\
\midrule
LibreOffice & 35 & 85.7\% & 0.62 \\
Chrome & 15 & 100.0\% & N/A \\
VSCode & 12 & 100.0\% & 1.00 \\
Thunderbird & 11 & 90.9\% & 0.82 \\
GIMP & 8 & 100.0\% & 1.00 \\
OS & 19 & 100.0\% & 1.00 \\
\midrule
\textbf{Overall} & \textbf{100} & \textbf{94.0\%} & \textbf{0.84} \\
\bottomrule
\end{tabular}
\end{table}

\subsection{Verification Tool-Usage Details}
\label{sec:appendix_tool_usage}

Table~\ref{tab:tool_usage_pct} summarizes how the three IRA backbones acquire evidence. VM commands are the primary evidence source for the GPT backbones and still account for nearly half of Qwen3.6's calls because they directly expose files, configurations, and system state. GPT-5.5 and GPT-5.4 require only 2.34 and 2.53 tool invocations per task, whereas Qwen3.6 uses 7.31 and relies much more heavily on GUI tools. Thus, the GPT backbones tend to select a small number of decisive checks, while Qwen3.6 compensates with broader interactive exploration.

%% file: supplementary_sections/reward_agent_trajectory_appendix.tex
\noindent\textbf{Hidden document content: receipt unit price}\label{app:traj:receipt-details-docx}
\noindent I have an image of my receipt located in /home/user. Please take a closer look at it and record the unit price of a new set of pedal arms in a new doc file named Details.docx on the desktop. record the number only.\par\smallskip
\noindent\raBadge{case: receipt\_details\_docx}\raBadge{model: gpt-5.5}\raBadge{steps: 17}\raGoodBadge{Reward Agent: Success (1.00)}\raBadBadge{VLM: Partial Success (0.55)}\par\medskip
\begin{rainputbox}
\raBoxTitle{Screenshots provided to the reward model}
\raThought{\# VLM can see the start/end desktop state, but not hidden file, config, or application state.}
\begin{minipage}[t]{0.48\linewidth}
\centering
\includegraphics[width=\linewidth]{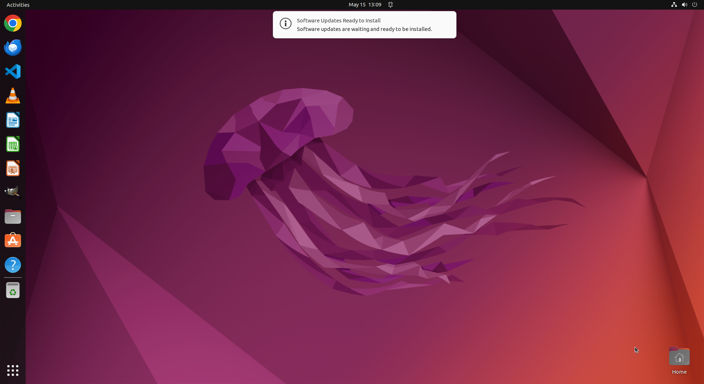}\\[-0.2ex]
{\scriptsize\ttfamily A. initial state}
\end{minipage}\hfill
\begin{minipage}[t]{0.48\linewidth}
\centering
\includegraphics[width=\linewidth]{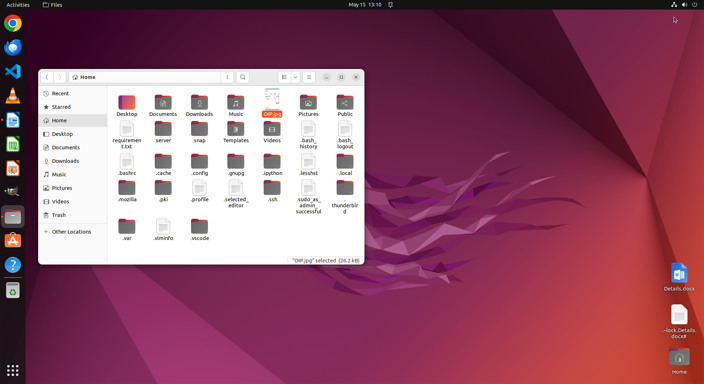}\\[-0.2ex]
{\scriptsize\ttfamily B. attempted end state}
\end{minipage}
\end{rainputbox}

\begin{ratoolbox}
\raStepHeader{01}{execute\_vm\_command}
\raCmd{\$ find /home/user -maxdepth 2 -type f -printf '\%p ' | sort}
\raThought{\# thought: I need to verify both the created document and the source receipt image. The screenshot suggests a}
\raThought{Details.docx file exists on the desktop, but the content and correctness of the recorded unit price are hidden. I will}
\raThought{first list the relevant files in /home/user and on the Desktop to identify the receipt image and the produced}
\raThought{document.}
\raObs{=> /home/user/.bash\_history /home/user/.bash\_logout /home/user/.bashrc}
\raObs{/home/user/.cache/event-sound-cache.tdb.9ec62d02362e449e91284972203df915.x86\_64-pc-linux-gnu...}
\smallskip
\raStepHeader{02}{get\_vm\_file}
\raCmd{\$ get\_vm\_file /home/user/Desktop/Details.docx -> Details.docx}
\raThought{\# thought: The file listing confirms the relevant source image is likely /home/user/OIP.jpg and that}
\raThought{/home/user/Desktop/Details.docx exists. Next I need to inspect the DOCX contents to see what number was recorded. I}
\raThought{will download the document for structured Word verification.}
\raObs{=> cache/e76c3496-4150-40c5-a306-344519ecf87c/Details.docx}
\smallskip
\raStepHeader{03}{checkwordfile}
\raCmd{\$ checkwordfile Details.docx --check 'Document contains the text "1450" and no other substantive text besides this}
\raCmd{number.'}
\raThought{\# thought: Now I have the produced Details.docx on the host. The task requires it to contain only the unit price}
\raThought{number for a new set of pedal arms. I will inspect the Word file text content.}
\raObs{=> \{"passed": false, "reason": "The document does not contain the number \textbackslash{}"1450\textbackslash{}".", "llm\_code\_path":...}
\smallskip
\raStepHeader{04}{execute\_vm\_command}
\raCmd{\$ python3 extract word/document.xml text from Details.docx}
\raThought{\# thought: The targeted Word check failed for the guessed value, so I should not rely on that assumption. I need to}
\raThought{determine the actual text in Details.docx and independently inspect the receipt image to identify the correct unit}
\raThought{price. I will extract the document text directly.}
\raObs{=> 15}
\smallskip
\raStepHeader{05}{get\_vm\_file}
\raCmd{\$ get\_vm\_file /home/user/OIP.jpg -> OIP.jpg}
\raThought{\# thought: The document contains only "15", which satisfies the "number only" formatting if the correct unit price is}
\raThought{15. I must verify the receipt image itself to determine the unit price of a new set of pedal arms. I will retrieve the}
\raThought{source image from /home/user/OIP.jpg for inspection.}
\raObs{=> cache/e76c3496-4150-40c5-a306-344519ecf87c/OIP.jpg}
\smallskip
\raStepHeader{06}{execute\_vm\_command}
\raCmd{\$ inspect /home/user/OIP.jpg metadata and OCR availability}
\raThought{\# thought: I need to read the receipt image. Since OCR may be available locally, I will first use read-only CLI image}
\raThought{metadata/OCR attempts to inspect the image contents without altering anything.}
\raObs{=> (474, 670) RGB python packages check pytesseract no No module named 'pytesseract'}
\smallskip
\end{ratoolbox}

\begin{raguibox}
\begin{minipage}[t]{0.43\linewidth}
\vspace{0pt}
\raGuiStepHeader{07}{computer}
\raGuiCmd{computer[double\_click] at [390, 172]}
\raGuiThought{\# thought: OCR is not available through the initial command, so I}
\raGuiThought{need a visual inspection path. The file manager in the current}
\raGuiThought{screenshot shows OIP.jpg selected, so I will open it in the image}
\raGuiThought{viewer to read the receipt.}
\raGuiObs{=> Computer action executed: double\_click}
\end{minipage}\hfill
\begin{minipage}[t]{0.53\linewidth}
\vspace{0pt}
\centering
\raComputerPointImage{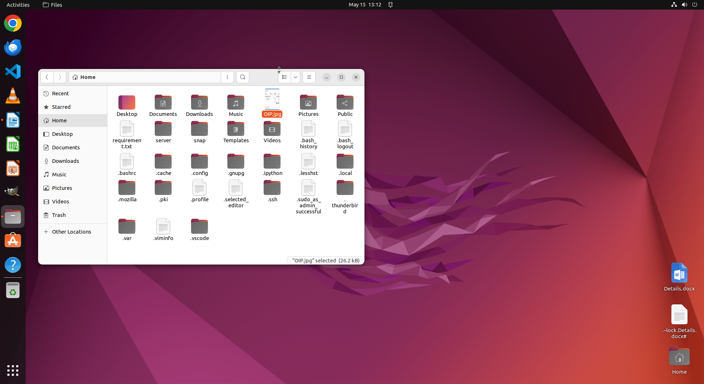}{390}{172}{double\_click}\\[-0.2ex]
{\scriptsize\ttfamily receipt\_details\_docx\_step-007-after.png}
\end{minipage}
\end{raguibox}

\begin{raguibox}
\begin{minipage}[t]{0.43\linewidth}
\vspace{0pt}
\raGuiStepHeader{08}{computer}
\raGuiCmd{computer[double\_click] at [370, 277]}
\raGuiThought{\# thought: The previous double-click did not open the image; the}
\raGuiThought{screenshot still shows the file manager with OIP.jpg selected. I}
\raGuiThought{likely clicked above the icon area. I will double-click the selected}
\raGuiThought{OIP.jpg icon more centrally to open it for visual inspection.}
\raGuiObs{=> Computer action executed: double\_click}
\end{minipage}\hfill
\begin{minipage}[t]{0.53\linewidth}
\vspace{0pt}
\centering
\raComputerPointImage{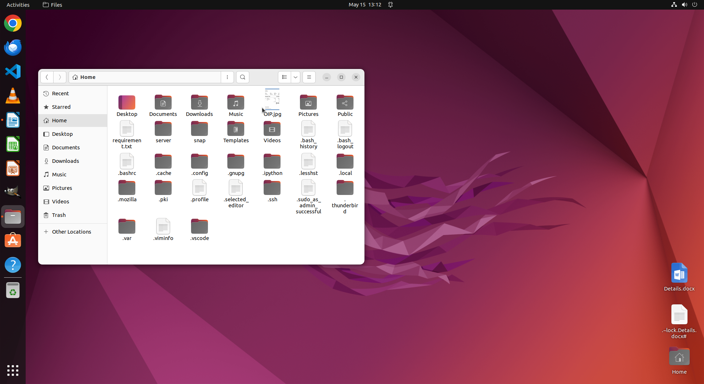}{370}{277}{double\_click}\\[-0.2ex]
{\scriptsize\ttfamily receipt\_details\_docx\_step-008-after.png}
\end{minipage}
\end{raguibox}

\begin{raguibox}
\begin{minipage}[t]{0.43\linewidth}
\vspace{0pt}
\raGuiStepHeader{09}{computer}
\raGuiCmd{computer[key] ENTER}
\raGuiThought{\# thought: The image appears selected but still not opened. Since}
\raGuiThought{selecting a file and pressing Enter is a standard way to open it, I}
\raGuiThought{will use Enter to open the selected OIP.jpg and inspect the receipt}
\raGuiThought{visually.}
\raGuiObs{=> Computer action executed: key}
\end{minipage}\hfill
\begin{minipage}[t]{0.53\linewidth}
\vspace{0pt}
\centering
\raComputerLabelImage{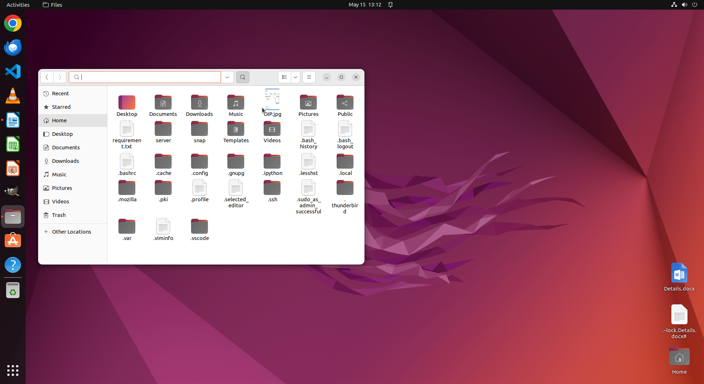}{key: ENTER}\\[-0.2ex]
{\scriptsize\ttfamily receipt\_details\_docx\_step-009-after.png}
\end{minipage}
\end{raguibox}

\begin{raguibox}
\begin{minipage}[t]{0.43\linewidth}
\vspace{0pt}
\raGuiStepHeader{10}{computer}
\raGuiCmd{computer[key] ESC}
\raGuiThought{\# thought: The file did not open because my earlier coordinates were}
\raGuiThought{off; the OIP.jpg icon is near the upper middle of the file manager,}
\raGuiThought{and the search/location field is now focused. I will press Escape to}
\raGuiThought{leave the field, then select/open the correct OIP.jpg icon area.}
\raGuiObs{=> Computer action executed: key}
\end{minipage}\hfill
\begin{minipage}[t]{0.53\linewidth}
\vspace{0pt}
\centering
\raComputerLabelImage{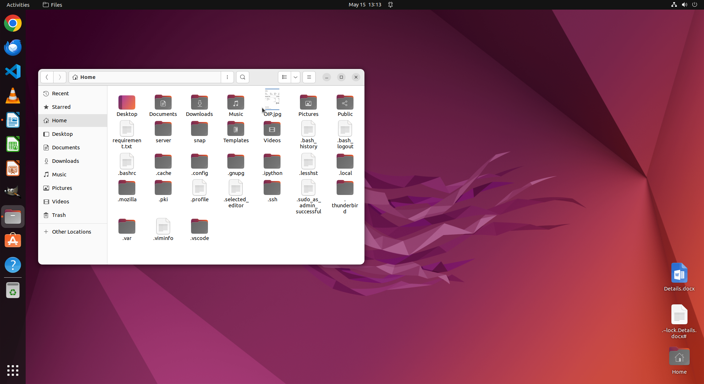}{key: ESC}\\[-0.2ex]
{\scriptsize\ttfamily receipt\_details\_docx\_step-010-after.png}
\end{minipage}
\end{raguibox}

\begin{ratoolbox}
\raStepHeader{11}{execute\_vm\_command}
\raCmd{\$ which xdg-open eog gimp display convert python3; ls target files}
\raThought{\# thought: I need more reliable visual reading of the image. Instead of struggling with GUI coordinates, I can use}
\raThought{command-line image processing to enlarge/convert the image, perhaps extract embedded text using available tools like}
\raThought{ImageMagick or save a zoomed version, then open that. First I will list whether image display tools and OCR binaries}
\raThought{are installed.}
\raObs{=> /usr/bin/xdg-open /usr/bin/eog /usr/bin/gimp /usr/bin/python3 -rw-rw-r-- 1 user user 4245 May 15 13:09}
\raObs{/home/user/Desktop/Details.docx -rw-r--r-- 1 user user 26173 May 15 13:08...}
\smallskip
\raStepHeader{12'}{invalid}
\raThought{\# thought: The document content is known, but I still need strict evidence for the receipt value. The file manager}
\raThought{coordinates have been unreliable, so I will inspect the accessibility tree to locate the OIP.jpg item and understand}
\raThought{the current UI structure before using further GUI interaction.}
\raObs{=> Model returned text without calling any tool}
\smallskip
\raStepHeader{12}{execute\_vm\_command}
\raCmd{\$ eog /home/user/OIP.jpg >/tmp/eog\_receipt.log 2>\&1 \&}
\raThought{\# thought: eog is available, so I can open the image directly via read-only xdg/eog command for visual inspection.}
\raThought{Although launching a viewer changes UI state, it is acceptable for verification and does not alter the image.}
\smallskip
\end{ratoolbox}

\begin{raguibox}
\begin{minipage}[t]{0.43\linewidth}
\vspace{0pt}
\raGuiStepHeader{13}{computer}
\raGuiCmd{computer[wait] 1s}
\raGuiThought{\# thought: I launched the image viewer in the background. I need to}
\raGuiThought{wait briefly for it to appear and then use the updated screenshot as}
\raGuiThought{visual evidence to read the receipt.}
\raGuiObs{=> Waited 1 seconds}
\end{minipage}\hfill
\begin{minipage}[t]{0.53\linewidth}
\vspace{0pt}
\centering
\raComputerLabelImage{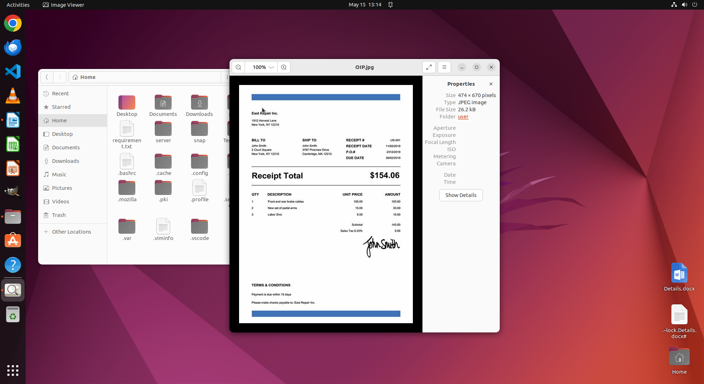}{wait: 1s}\\[-0.2ex]
{\scriptsize\ttfamily receipt\_details\_docx\_step-013-after.png}
\end{minipage}
\end{raguibox}

\begin{raguibox}
\begin{minipage}[t]{0.43\linewidth}
\vspace{0pt}
\raGuiStepHeader{14}{computer}
\raGuiCmd{computer[key] CTRL++}
\raGuiThought{\# thought: The receipt image is now open and visible, but the item}
\raGuiThought{table text is still small. I can see the target row appears in the}
\raGuiThought{table, but I need higher confidence about the unit price}
\raGuiThought{format/value. I will zoom in on the image viewer so the row for "new}
\raGuiThought{set of pedal arms" can be read more clearly.}
\raGuiObs{=> Computer action executed: key}
\end{minipage}\hfill
\begin{minipage}[t]{0.53\linewidth}
\vspace{0pt}
\centering
\raComputerLabelImage{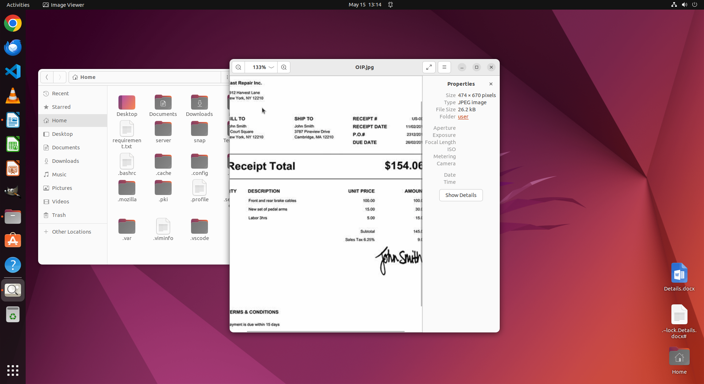}{key: CTRL++}\\[-0.2ex]
{\scriptsize\ttfamily receipt\_details\_docx\_step-014-after.png}
\end{minipage}
\end{raguibox}

\begin{rafinalbox}
\raStepHeader{15}{final\_answer}
\raCmd{reward=1 verdict=Success}
\raThought{\# thought: The evidence now shows the source receipt and the produced document. The receipt row for "New set of pedal}
\raThought{arms" shows a unit price of 15.00, and the created /home/user/Desktop/Details.docx contains only "15", which is the}
\raThought{requested number without currency or extra text. All critical criteria are satisfied.}
\raObs{=> Evaluation completed}
\raFinalText{The required document exists at /home/user/Desktop/Details.docx. Direct extraction of the DOCX text showed it contains}
\raFinalText{only the text "15". Visual inspection of the receipt image /home/user/OIP.jpg shows the item "New set of pedal arms"}
\raFinalText{with unit price 15.00, so recording the number only as "15" satisfies the task.}
\end{rafinalbox}

\noindent\textbf{Hidden browser state: startup URL removed}\label{app:traj:chrome-funbrain-startup}
\noindent On my surface pro whenever I launch Chrome it always opens "funbrain.com." I don't want this. I cleared my cache but it still happens. What should I do?\par\smallskip
\noindent\raBadge{case: chrome\_funbrain\_startup}\raBadge{model: gpt-5.5}\raBadge{steps: 10}\raGoodBadge{Reward Agent: Success (1.00)}\raBadBadge{VLM: Failure (0.10)}\par\medskip
\begin{rainputbox}
\raBoxTitle{Screenshots provided to the reward model}
\raThought{\# VLM can see the start/end desktop state, but not hidden file, config, or application state.}
\begin{minipage}[t]{0.48\linewidth}
\centering
\includegraphics[width=\linewidth]{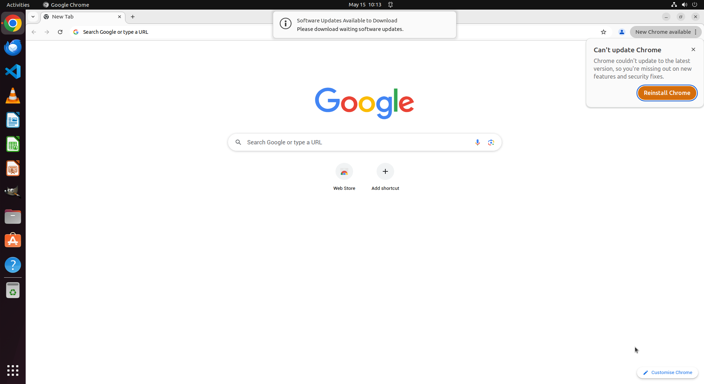}\\[-0.2ex]
{\scriptsize\ttfamily A. initial state}
\end{minipage}\hfill
\begin{minipage}[t]{0.48\linewidth}
\centering
\includegraphics[width=\linewidth]{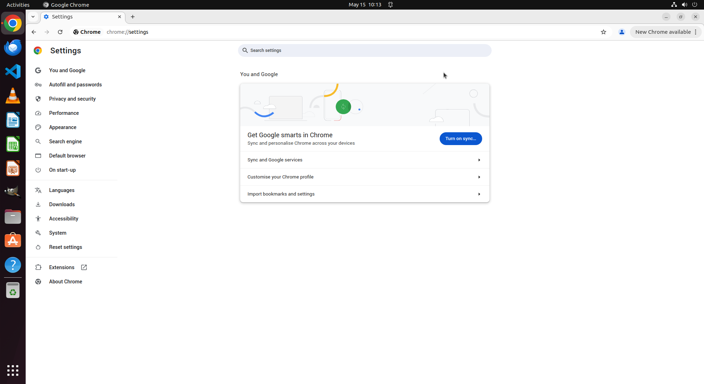}\\[-0.2ex]
{\scriptsize\ttfamily B. attempted end state}
\end{minipage}
\end{rainputbox}

\begin{ratoolbox}
\raStepHeader{01}{execute\_vm\_command}
\raCmd{\$ ls -la /home/oai/.config/google-chrome \&\& find /home/oai/.config/google-chrome -maxdepth 2 -type f -name}
\raCmd{'Preferences' -o -name 'Secure Preferences'}
\raThought{\# thought: I need to verify Chrome's actual startup configuration, because the task is about Chrome opening}
\raThought{funbrain.com on launch and the relevant completion condition is hidden in browser preferences rather than visible on}
\raThought{the main Settings page. I will inspect the Chrome profile configuration files to see whether funbrain.com remains}
\raThought{configured as a startup page or related homepage setting.}
\raObs{=> \{'output': '', 'error': "ls: cannot access '/home/oai/.config/google-chrome': No such file or directory\textbackslash{}n"\}}
\smallskip
\raStepHeader{02}{execute\_vm\_command}
\raCmd{\$ echo \$HOME; ls -la /home; find /home -maxdepth 4 -type f \textbackslash{}( -name 'Preferences' -o -name 'Secure Preferences' \textbackslash{})}
\raCmd{2>/dev/null}
\raThought{\# thought: The expected Chrome profile path was not present, so I cannot conclude from that location. I need to}
\raThought{identify the actual user/config location and then inspect the relevant Chrome/Chromium preference files without}
\raThought{assuming the path.}
\raObs{=> /home/user total 12 drwxr-xr-x 3 root root 4096 Dec 17 2023 . drwxr-xr-x 20 root root 4096 Oct 18 2024 ..}
\raObs{drwxr-x--- 23 user user 4096 Aug 5 2025 user /home/user/.config/Code/Preferences}
\smallskip
\raStepHeader{03}{execute\_vm\_command}
\raCmd{\$ ls -la /home/user/.config; ps -eo pid,args | sed -n '1,200p'}
\raThought{\# thought: The home directory is /home/user, and the shallow search did not find Chrome Preferences. Since Chrome is}
\raThought{visibly running, its profile may be in a nonstandard directory or deeper path. I will list the relevant configuration}
\raThought{directories and running command line to locate the active user data directory.}
\raObs{=> total 120 drwx------ 23 user user 4096 Aug 5 2025 . drwxr-x--- 23 user user 4096 Aug 5 2025 .. drwxrwxr-x 2 user}
\raObs{user 4096 Mar 17 2024 autostart drwx------ 19 user user 4096 Aug 5 2025 Code...}
\smallskip
\raStepHeader{04}{execute\_vm\_command}
\raCmd{\$ find /home/user/.config/google-chrome -maxdepth 3 -type f \textbackslash{}( -name 'Preferences' -o -name 'Secure Preferences' -o}
\raCmd{-name 'Local State' \textbackslash{}) -print -exec python3 - <<'PY' \{\} \textbackslash{}; import json,sys,os p=sys.argv[1] print(' ---',p,'---') try:}
\raCmd{data=json.load(open(p)) except Exception as e: print('ERR',e); sys.exit() \# Print full top-level keys and selected}
\raCmd{relevant sections if present print('top\_keys=', sorted(data.keys())) for key in ['session','homepage','homepage\_is\_new}
\raCmd{tabpage','startup\_urls','browser','profile','extensions','default\_search\_provider','restore\_on\_startup','restore\_on\_st}
\raCmd{artup\_migrated']: if key in data: print(key, json.dumps(data[key], indent=2)[:12000]) PY}
\raThought{\# thought: I see the google-chrome config directory exists under /home/user. I need the preference details, especially}
\raThought{startup URLs, restore behavior, homepage, and possibly extension settings that could force funbrain.com. I will print}
\raThought{the relevant preference files in a structured way.}
\raObs{=> /home/user/.config/google-chrome/Local State --- /home/user/.config/google-chrome/Local State --- top\_keys=}
\raObs{['accessibility', 'autofill', 'background\_tracing', 'breadcrumbs', 'browser',...}
\smallskip
\raStepHeader{05}{execute\_vm\_command}
\raCmd{\$ python3 - <<'PY' import json, os}
\raCmd{paths=['/home/user/.config/google-chrome/Default/Preferences','/home/user/.config/google-chrome/Default/Secure}
\raCmd{Preferences'] for p in paths: print(' ---',p,'exists',os.path.exists(p),'---') if not os.path.exists(p): continue}
\raCmd{txt=open(p,errors='replace').read() print('contains funbrain:', 'funbrain' in txt.lower()) try: data=json.loads(txt)}
\raCmd{except Exception as e: print('json error',e); continue print('top\_keys=', sorted(data.keys())) for key in ['session','}
\raCmd{homepage','homepage\_is\_newtabpage','startup\_urls','browser','profile','extensions','default\_search\_provider','restore\_}
\raCmd{on\_startup','restore\_on\_startup\_migrated']: if key in data: out=json.dumps(data[key], indent=2, sort\_keys=True)}
\raCmd{print(' '+key+':') print(out[:20000]) PY}
\raThought{\# thought: The previous command only displayed Local State and did not successfully show the Default profile}
\raThought{preference contents, which are the main evidence. I need to inspect the Default profile Preferences and Secure}
\raThought{Preferences directly for startup/homepage settings and any occurrences of funbrain.com.}
\raObs{=> --- /home/user/.config/google-chrome/Default/Preferences exists True --- contains funbrain: False top\_keys=}
\raObs{['NewTabPage', 'accessibility', 'account\_tracker\_service\_last\_update',...}
\smallskip
\raStepHeader{06}{execute\_vm\_command}
\raCmd{\$ printf ' -- Chrome policy dirs -- '; for d in /etc/opt/chrome /etc/chromium /etc/google-chrome}
\raCmd{/usr/share/google-chrome; do echo "DIR \$d"; find "\$d" -maxdepth 4 -type f 2>/dev/null | sort; done; printf ' --}
\raCmd{desktop entries containing chrome/google -- '; find /usr/share/applications /home/user/.local/share/applications}
\raCmd{/home/user/Desktop -maxdepth 2 -type f 2>/dev/null | sort | while read f; do case "\$f" in}
\raCmd{*chrome*|*Chrome*|*google*|*Google*) echo "FILE \$f"; sed -n '1,220p' "\$f"; echo; esac; done}
\raThought{\# thought: The Default Preferences show no occurrence of funbrain and no printed startup URL settings, which suggests}
\raThought{the startup page may have been removed. However, launch behavior can also be controlled by Chrome policies or desktop}
\raThought{shortcuts/command-line arguments, so I need to inspect those system-backed sources before judging success.}
\raObs{=> -- Chrome policy dirs -- DIR /etc/opt/chrome DIR /etc/chromium DIR /etc/google-chrome DIR /usr/share/google-chrome}
\raObs{-- desktop entries containing chrome/google -- FILE...}
\smallskip
\raStepHeader{07}{execute\_vm\_command}
\raCmd{\$ python3 - <<'PY' import json, os, re base='/home/user/.config/google-chrome/Default' for name in}
\raCmd{['Preferences','Secure Preferences']: p=os.path.join(base,name) print(' ---',p,'---') data=json.load(open(p)) def}
\raCmd{walk(x,path=''): if isinstance(x, dict): for k,v in x.items(): np=f'\{path\}.\{k\}' if path else k s=k.lower() if any(term}
\raCmd{in s for term in ['startup','restore','homepage','home\_page','session']): print(np,'=',repr(v)[:1000]) walk(v,np) elif}
\raCmd{isinstance(x, list): for i,v in enumerate(x): walk(v,f'\{path\}[\{i\}]') else: if isinstance(x,str): lx=x.lower() if}
\raCmd{any(term in lx for term in ['funbrain','http://','https://','chrome://newtab','chrome://settings']):}
\raCmd{print(path,'=',repr(x)[:1000]) walk(data) PY}
\raThought{\# thought: The main profile and desktop launcher do not show funbrain.com, but I still need to verify the exact Chrome}
\raThought{startup-related preference values rather than infer from missing printed sections. I will recursively enumerate}
\raThought{keys/values containing startup, restore, homepage, and any URL-like occurrences in the active Chrome profile.}
\raObs{=> --- /home/user/.config/google-chrome/Default/Preferences ---}
\raObs{extensions.settings.ahfgeienlihckogmohjhadlkjgocpleb.manifest.app.launch.web\_url =}
\raObs{'https://chrome.google.com/webstore'...}
\smallskip
\raStepHeader{08}{execute\_vm\_command}
\raCmd{\$ grep -RIl --exclude='*.log' --exclude='*.ldb' --exclude='*.sqlite*' --exclude='History' --exclude='Cookies'}
\raCmd{--exclude='Cache\_Data' 'funbrain' /home/user/.config/google-chrome /home/user/.local/share/applications}
\raCmd{/home/user/Desktop /usr/share/applications /etc/opt/chrome /etc/chromium 2>/dev/null || true}
\raThought{\# thought: The active profile's actual preference values do not contain funbrain.com or startup URL keys, while only}
\raThought{MAC hashes remain. To rule out another persistent source such as history/session files, extension state, or shortcuts}
\raThought{containing funbrain.com, I will search the Chrome config and common launch locations for any occurrence of "funbrain".}
\smallskip
\end{ratoolbox}

\begin{rafinalbox}
\raStepHeader{09}{final\_answer}
\raCmd{reward=1 verdict=Success}
\raThought{\# thought: The task requires the final Chrome launch/startup configuration to no longer open funbrain.com. Although}
\raThought{the before/after screenshots look visually unchanged on the general Chrome Settings page, the decisive state is hidden}
\raThought{in Chrome's profile, policy, and launcher configuration. I verified the active Chrome profile under}
\raThought{/home/user/.config/google-chrome/Default: Preferences and Secure Preferences contain no funbrain.com, no session}
\raThought{startup\_urls setting, and no homepage setting pointing to funbrain; policy directories and the Google Chrome desktop}
\raThought{launcher also do not contain funbrain.com. There are only built-in extensions, with no apparent extension override.}
\raThought{Therefore the persistent sources that would make Chrome open funbrain.com on launch appear removed/absent, satisfying}
\raThought{the objective.}
\raObs{=> Evaluation completed}
\raFinalText{Final state verifies the required end condition: Chrome is no longer configured to launch funbrain.com. The active}
\raFinalText{Chrome profile Preferences contain no occurrence of funbrain.com and no startup\_urls/homepage setting pointing to it;}
\raFinalText{Secure Preferences also contain no occurrence. Chrome policy/launcher locations inspected did not contain}
\raFinalText{funbrain.com, and the installed extensions shown in Preferences are only built-in components with no override. The}
\raFinalText{screenshots themselves are visually unchanged on the main Settings page, but the hidden configuration evidence}
\raFinalText{confirms the unwanted startup page is not configured.}
\end{rafinalbox}

\noindent\textbf{Persistence check: VLC minimal interface}\label{app:traj:vlc-minimal-persistence}
\noindent Enable VLC Minimal Interface in window mode so the bottom playback controls are hidden, and make sure the setting persists after restarting VLC. I often multitask on my computer, and the persistent toolbar in VLC is very distracting.\par\smallskip
\noindent\raBadge{case: vlc\_minimal\_persistence}\raBadge{model: gpt-5.5}\raBadge{steps: 9}\raGoodBadge{Reward Agent: Success (1.00)}\raBadBadge{VLM: Partial Success (0.75)}\par\medskip
\begin{rainputbox}
\raBoxTitle{Screenshots provided to the reward model}
\raThought{\# VLM can see the start/end desktop state, but not hidden file, config, or application state.}
\begin{minipage}[t]{0.48\linewidth}
\centering
\includegraphics[width=\linewidth]{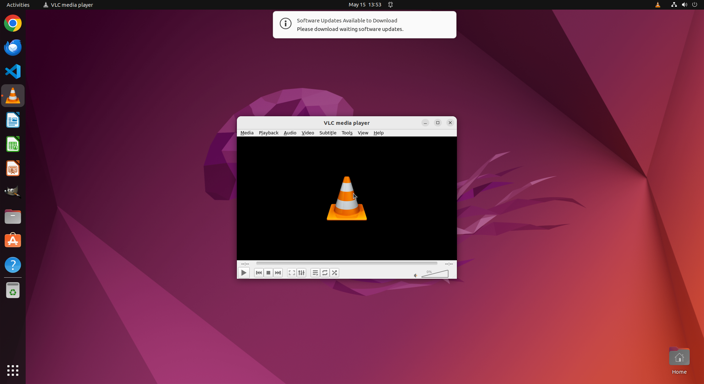}\\[-0.2ex]
{\scriptsize\ttfamily A. initial state}
\end{minipage}\hfill
\begin{minipage}[t]{0.48\linewidth}
\centering
\includegraphics[width=\linewidth]{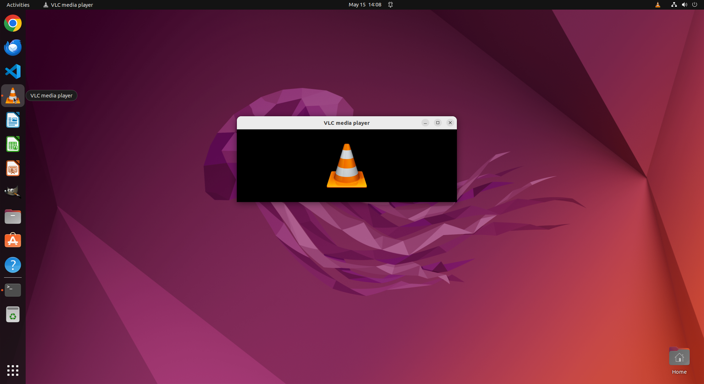}\\[-0.2ex]
{\scriptsize\ttfamily B. attempted end state}
\end{minipage}
\end{rainputbox}

\begin{ratoolbox}
\raStepHeader{01}{execute\_vm\_command}
\raCmd{\$ ls -la /home/oai/.config/vlc \&\& sed -n '1,260p' /home/oai/.config/vlc/vlcrc}
\raThought{\# thought: I need to verify both critical requirements: VLC is currently in windowed minimal interface with the bottom}
\raThought{controls hidden (visible in Image B), and the preference is persisted so it will remain after VLC restarts. The}
\raThought{visible screenshot strongly suggests the controls and menu are hidden, but persistence is a hidden configuration}
\raThought{criterion, so I will inspect VLC's configuration files rather than rely on appearance alone.}
\raObs{=> \{'output': '', 'error': "ls: cannot access '/home/oai/.config/vlc': No such file or directory\textbackslash{}n"\}}
\smallskip
\raStepHeader{02}{execute\_vm\_command}
\raCmd{\$ printf 'HOME=\%s ' "\$HOME"; ls -la /home; ls -la "\$HOME/.config"; if [ -d "\$HOME/.config/vlc" ]; then ls -la}
\raCmd{"\$HOME/.config/vlc"; fi}
\raThought{\# thought: The attempted config inspection used the wrong home path, as the directory `/home/oai/.config/vlc` does not}
\raThought{exist. I need to locate the actual user home and inspect the VLC configuration there, without assuming the username.}
\raObs{=> HOME=/home/user total 12 drwxr-xr-x 3 root root 4096 Dec 17 2023 . drwxr-xr-x 20 root root 4096 Oct 18 2024 ..}
\raObs{drwxr-x--- 23 user user 4096 Aug 5 2025 user total 120 drwx------ 23 user user...}
\smallskip
\raStepHeader{03}{execute\_vm\_command}
\raCmd{\$ cat /home/user/.config/vlc/vlc-qt-interface.conf}
\raThought{\# thought: The correct VLC config directory exists under `/home/user/.config/vlc`. To verify persistence, I need to}
\raThought{inspect the Qt interface configuration file where VLC stores window/interface state, especially whether minimal view}
\raThought{is enabled after restart.}
\raObs{=> [General] filedialog-path=@Variant(\textbackslash{}0\textbackslash{}0\textbackslash{}0\textbackslash{}x11\textbackslash{}0\textbackslash{}0\textbackslash{}0\textbackslash{}n/home/user)...}
\smallskip
\end{ratoolbox}

\begin{raguibox}
\begin{minipage}[t]{0.43\linewidth}
\vspace{0pt}
\raGuiStepHeader{04}{computer}
\raGuiCmd{computer[left\_click]}
\raGuiThought{\# thought: The Qt interface config confirms status bar is hidden,}
\raGuiThought{but it does not explicitly show the minimal-interface toggle or}
\raGuiThought{whether the bottom playback controls will remain hidden after a}
\raGuiThought{restart. Because persistence after restart is a central task}
\raGuiThought{requirement, I should verify it behaviorally by restarting VLC}
\raGuiThought{through the GUI and checking whether the window reopens in minimal}
\raGuiThought{interface. First I will close the current VLC window.}
\raGuiObs{=> Computer action executed: left\_click}
\end{minipage}\hfill
\begin{minipage}[t]{0.53\linewidth}
\vspace{0pt}
\centering
\raComputerLabelImage{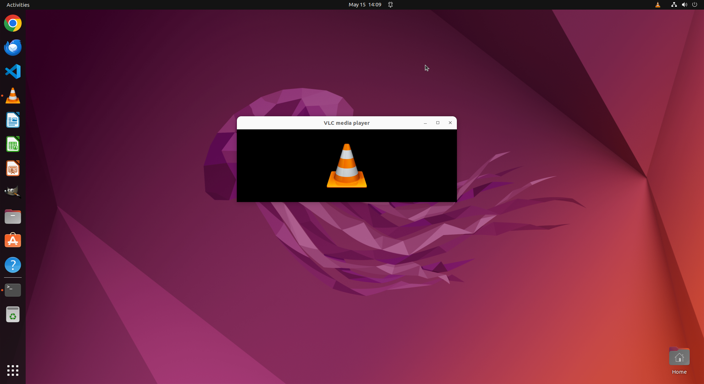}{left\_click}\\[-0.2ex]
{\scriptsize\ttfamily vlc\_minimal\_persistence\_step-004-after.png}
\end{minipage}
\end{raguibox}

\begin{raguibox}
\begin{minipage}[t]{0.43\linewidth}
\vspace{0pt}
\raGuiStepHeader{05}{computer}
\raGuiCmd{computer[left\_click]}
\raGuiThought{\# thought: The previous click did not close VLC; the window is still}
\raGuiThought{visible in minimal interface. I need to close it using the close}
\raGuiThought{button at the top-right of the VLC window so that I can restart VLC}
\raGuiThought{and verify persistence.}
\raGuiObs{=> Computer action executed: left\_click}
\end{minipage}\hfill
\begin{minipage}[t]{0.53\linewidth}
\vspace{0pt}
\centering
\raComputerLabelImage{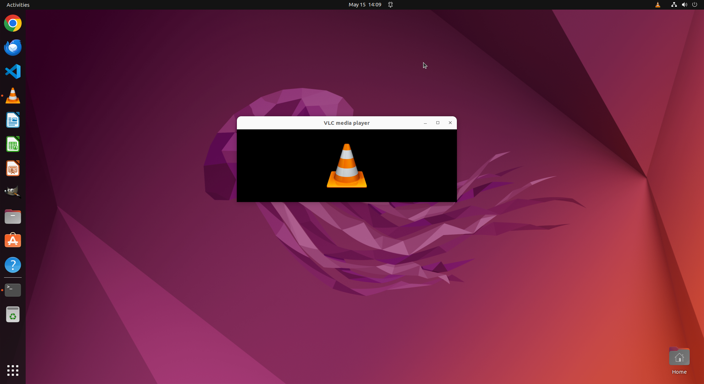}{left\_click}\\[-0.2ex]
{\scriptsize\ttfamily vlc\_minimal\_persistence\_step-005-after.png}
\end{minipage}
\end{raguibox}

\begin{raguibox}
\begin{minipage}[t]{0.43\linewidth}
\vspace{0pt}
\raGuiStepHeader{06}{computer}
\raGuiCmd{computer[left\_click]}
\raGuiThought{\# thought: I realize my previous close clicks used incorrect}
\raGuiThought{relative coordinates (too high on the screen); the VLC close button}
\raGuiThought{is near the window's top-right at about x 450, y 122 in the 768x432}
\raGuiThought{screenshot, which corresponds to roughly [586, 282]. I will click}
\raGuiThought{that actual close button now to restart VLC for persistence}
\raGuiThought{verification.}
\raGuiObs{=> Computer action executed: left\_click}
\end{minipage}\hfill
\begin{minipage}[t]{0.53\linewidth}
\vspace{0pt}
\centering
\raComputerPointImage{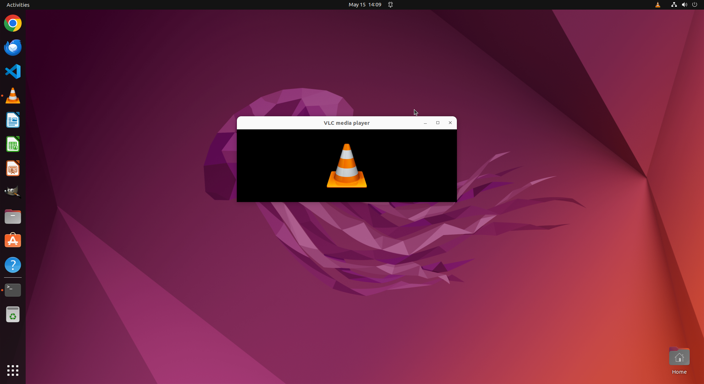}{586}{282}{left\_click}\\[-0.2ex]
{\scriptsize\ttfamily vlc\_minimal\_persistence\_step-006-after.png}
\end{minipage}
\end{raguibox}

\begin{ratoolbox}
\raStepHeader{07}{execute\_vm\_command}
\raCmd{\$ ps -ef | grep -i '[v]lc'; printf ' --- vlc qt interface --- '; cat /home/user/.config/vlc/vlc-qt-interface.conf;}
\raCmd{printf ' --- vlcrc lines mentioning qt/minimal/interface/controls/status/toolbar --- '; sed -n '1,2600p'}
\raCmd{/home/user/.config/vlc/vlcrc | awk 'BEGIN\{IGNORECASE=1\} /qt|minimal|interface|controls|status|toolbar/ \{print NR ":"}
\raCmd{\$0\}'}
\raThought{\# thought: The GUI close attempts have not closed the window, so instead of repeatedly clicking, I will use read-only}
\raThought{process inspection/config inspection to gather more evidence. I need to see whether VLC is running and inspect}
\raThought{relevant configuration entries in both vlcrc and qt config for interface settings that would indicate a persistent}
\raThought{minimal interface.}
\raObs{=> user 9678 1005 0 14:08 ? 00:00:00 /usr/bin/vlc --started-from-file user 10258 2168 0 14:10 ? 00:00:00 /bin/sh -c ps}
\raObs{-ef | grep -i '[v]lc'; printf '\textbackslash{}n--- vlc qt interface ---\textbackslash{}n'; cat...}
\smallskip
\end{ratoolbox}

\begin{rafinalbox}
\raStepHeader{08}{final\_answer}
\raCmd{reward=1 verdict=Success}
\raThought{\# thought: The visible final state shows VLC in a window with only the video area and title bar, with the menu bar and}
\raThought{bottom playback controls hidden, satisfying the visible minimal-interface requirement. The hidden persistence}
\raThought{requirement is also supported by VLC's configuration: `/home/user/.config/vlc/vlcrc` contains `qt-minimal-view=1`}
\raThought{under the `[qt]` section, which is the persistent setting to start VLC in minimal view. Therefore both critical}
\raThought{criteria are satisfied.}
\raObs{=> Evaluation completed}
\raFinalText{Success. Image B/current screenshots show VLC in windowed minimal interface with no bottom playback controls or menu}
\raFinalText{bar visible. The persistent setting is verified in `/home/user/.config/vlc/vlcrc`, where the Qt interface section}
\raFinalText{contains `qt-minimal-view=1`, indicating VLC will start in minimal view after restart. The task's visible and}
\raFinalText{persistence requirements are both satisfied.}
\end{rafinalbox}

\noindent\textbf{Cross-app evidence: tuition reminder recipients}\label{app:traj:thunderbird-tuition-recipients}
\noindent I've drafted an e-mail reminder for those who haven't paid tuition. Please help me to check out their e-mails from the payment record and add to the receiver field.\par\smallskip
\noindent\raBadge{case: thunderbird\_tuition\_recipients}\raBadge{model: gpt-5.5}\raBadge{steps: 7}\raGoodBadge{Reward Agent: Success (1.00)}\raBadBadge{VLM: Partial Success (0.75)}\par\medskip
\begin{rainputbox}
\raBoxTitle{Screenshots provided to the reward model}
\raThought{\# VLM can see the start/end desktop state, but not hidden file, config, or application state.}
\begin{minipage}[t]{0.48\linewidth}
\centering
\includegraphics[width=\linewidth]{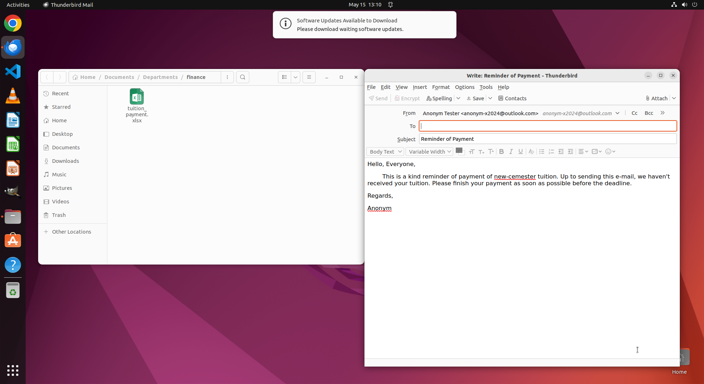}\\[-0.2ex]
{\scriptsize\ttfamily A. initial state}
\end{minipage}\hfill
\begin{minipage}[t]{0.48\linewidth}
\centering
\includegraphics[width=\linewidth]{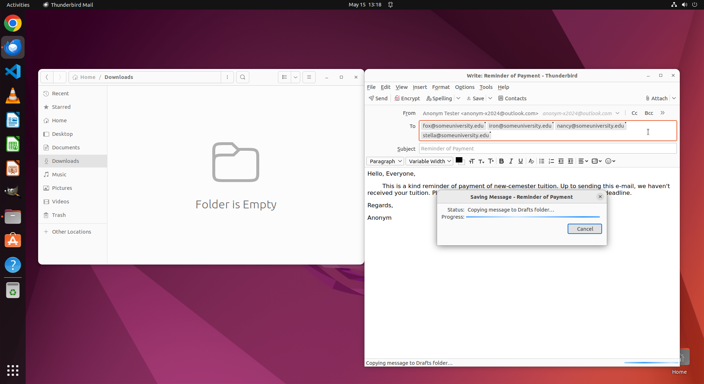}\\[-0.2ex]
{\scriptsize\ttfamily B. attempted end state}
\end{minipage}
\end{rainputbox}

\begin{ratoolbox}
\raStepHeader{01}{execute\_vm\_command}
\raCmd{\$ find /home -maxdepth 4 -type f | sed 's\#\textasciicircum{}\#/\#\#' | sort}
\raThought{\# thought: I need to verify the required final state: the Thunderbird draft's recipient field should contain the email}
\raThought{addresses of people who have not paid tuition, as determined from the payment record. The screenshots show the compose}
\raThought{window and a receiver field, but they do not reveal the payment record or allow reliable extraction of all recipients.}
\raThought{I will inspect the filesystem for the payment record or related files so I can compare the draft recipients against}
\raThought{the source data.}
\raObs{=> //home/user/.bash\_history //home/user/.bash\_logout //home/user/.bashrc //home/user/.cache/babl/babl-fishes}
\raObs{//home/user/.cache/dconf/user...}
\smallskip
\raStepHeader{02}{execute\_vm\_command}
\raCmd{\$ ls -laR /home/user/Documents /home/user/Downloads /home/user/Desktop 2>\&1}
\raThought{\# thought: The broad file listing did not show the tuition payment spreadsheet even though Image A showed it under}
\raThought{Documents/Departments/Finance. I need a focused directory listing to determine whether the payment record still exists}
\raThought{and where; if it is missing, that is important evidence because the task depended on it, but I need not assume absence}
\raThought{from the broad listing alone.}
\raObs{=> /home/user/Desktop: total 8 drwxr-xr-x 2 user user 4096 Mar 15 2024 . drwxr-x--- 23 user user 4096 May 15 13:09 ..}
\raObs{/home/user/Documents: total 12 drwxr-xr-x 3 user user 4096 May 15 13:09 ....}
\smallskip
\raStepHeader{03}{execute\_vm\_command}
\raCmd{\$ python3 - <<'PY' from openpyxl import load\_workbook}
\raCmd{path='/home/user/Documents/Departments/finance/tuition\_payment.xlsx' wb=load\_workbook(path, data\_only=True) for ws in}
\raCmd{wb.worksheets: print('SHEET', ws.title) for row in ws.iter\_rows(values\_only=True): print('\textbackslash{}t'.join('' if v is None}
\raCmd{else str(v) for v in row)) PY}
\raThought{\# thought: Now I have confirmed the payment record exists at}
\raThought{/home/user/Documents/Departments/finance/tuition\_payment.xlsx. I need to inspect its contents to identify exactly}
\raThought{which students have not paid and their email addresses, then compare that expected set with the Thunderbird recipient}
\raThought{field. A read-only Python command can dump the workbook contents.}
\raObs{=> \{'output': '', 'error': 'Traceback (most recent call last):\textbackslash{}n File "<stdin>", line 1, in}
\raObs{<module>\textbackslash{}nModuleNotFoundError: No module named \textbackslash{}'openpyxl\textbackslash{}'\textbackslash{}n'\}}
\smallskip
\raStepHeader{04}{execute\_vm\_command}
\raCmd{\$ python3 - <<'PY' import zipfile, xml.etree.ElementTree as ET}
\raCmd{path='/home/user/Documents/Departments/finance/tuition\_payment.xlsx' with zipfile.ZipFile(path) as z: print('FILES')}
\raCmd{for n in z.namelist(): print(n) ns=\{'a':'http://schemas.openxmlformats.org/spreadsheetml/2006/main'\} shared=[] if}
\raCmd{'xl/sharedStrings.xml' in z.namelist(): root=ET.fromstring(z.read('xl/sharedStrings.xml')) for si in}
\raCmd{root.findall('a:si', ns): texts=[] for t in si.iter('\{http://schemas.openxmlformats.org/spreadsheetml/2006/main\}t'):}
\raCmd{texts.append(t.text or '') shared.append(''.join(texts)) print('SHARED\_STRINGS') for i,s in enumerate(shared):}
\raCmd{print(i, repr(s)) \# workbook sheets and relationships sheets=[n for n in z.namelist() if}
\raCmd{n.startswith('xl/worksheets/sheet') and n.endswith('.xml')] print('SHEETS') for sheet in sheets: print('---', sheet)}
\raCmd{root=ET.fromstring(z.read(sheet)) for row in root.findall('.//a:sheetData/a:row', ns): vals=[] for c in}
\raCmd{row.findall('a:c', ns): ref=c.attrib.get('r') typ=c.attrib.get('t') v=c.find('a:v', ns) val='' if v is None or v.text}
\raCmd{is None else v.text if typ=='s' and val!='': val=shared[int(val)] elif typ=='inlineStr': texts=[] for t in}
\raCmd{c.iter('\{http://schemas.openxmlformats.org/spreadsheetml/2006/main\}t'): texts.append(t.text or '') val=''.join(texts)}
\raCmd{vals.append(f'\{ref\}=\{val\}') print(' | '.join(vals)) PY}
\raThought{\# thought: The openpyxl library is unavailable, so I need another read-only way to inspect the xlsx. Since an .xlsx is}
\raThought{a ZIP archive of XML files, I can list and parse its XML contents directly to recover the table values.}
\raObs{=> FILES \_rels/.rels xl/\_rels/workbook.xml.rels xl/workbook.xml xl/theme/theme1.xml xl/styles.xml}
\raObs{xl/worksheets/sheet1.xml xl/sharedStrings.xml docProps/core.xml docProps/app.xml...}
\smallskip
\raStepHeader{05}{get\_accessibility\_tree}
\raCmd{\{\}}
\raThought{\# thought: The spreadsheet shows the unpaid students are Fox, Iron, Nancy, and Stella, with emails}
\raThought{fox@someuniversity.edu, iron@someuniversity.edu, nancy@someuniversity.edu, and stella@someuniversity.edu. Image B}
\raThought{visibly appears to contain those four addresses in the Thunderbird To field, but I want stronger structured evidence}
\raThought{for the compose window's recipient field before final judgment. The accessibility tree should expose the exact text in}
\raThought{the active UI.}
\raObs{=> <desktop-frame xmlns:st="https://accessibility.ubuntu.example.org/ns/state"}
\raObs{xmlns:attr="https://accessibility.ubuntu.example.org/ns/attributes"...}
\smallskip
\end{ratoolbox}

\begin{rafinalbox}
\raStepHeader{06}{final\_answer}
\raCmd{reward=1 verdict=Success}
\raThought{\# thought: I have verified the critical final-state requirements. The payment record lists "No" for Fox, Iron, Nancy,}
\raThought{and Stella, whose emails are fox@someuniversity.edu, iron@someuniversity.edu, nancy@someuniversity.edu, and}
\raThought{stella@someuniversity.edu. The Thunderbird compose window's To field contains exactly those four address pills and no}
\raThought{additional recipient pills. Although there is a draft-save error dialog, saving/sending was not part of the task, and}
\raThought{the required receiver field has been correctly populated.}
\raObs{=> Evaluation completed}
\raFinalText{The task was to find the unpaid students' emails from the payment record and add them to the email receiver field.}
\raFinalText{Inspecting the spreadsheet showed unpaid entries for Fox, Iron, Nancy, and Stella, with emails fox@someuniversity.edu,}
\raFinalText{iron@someuniversity.edu, nancy@someuniversity.edu, and stella@someuniversity.edu. The Thunderbird compose window's To}
\raFinalText{field contains exactly these four recipient address pills. No extra or missing recipients were found. The visible}
\raFinalText{draft-save error is not relevant to the requested receiver-field update.}
\end{rafinalbox}

\noindent\textbf{Spreadsheet structure beats misleading pixels}\label{app:traj:calc-old-id-padding}
\noindent I would like to copy all the numbers in the 'Old ID' column to the 'New 7 Digit Id' column, and pad them with zeros in front, to fill them up to seven digits. Finish the work and don't touch irrelevant regions, even if they are blank.\par\smallskip
\noindent\raBadge{case: calc\_old\_id\_padding}\raBadge{model: gpt-5.5}\raBadge{steps: 4}\raGoodBadge{Reward Agent: Success (1.00)}\raBadBadge{VLM: Failure (0.00)}\par\medskip
\begin{rainputbox}
\raBoxTitle{Screenshots provided to the reward model}
\raThought{\# VLM can see the start/end desktop state, but not hidden file, config, or application state.}
\begin{minipage}[t]{0.48\linewidth}
\centering
\includegraphics[width=\linewidth]{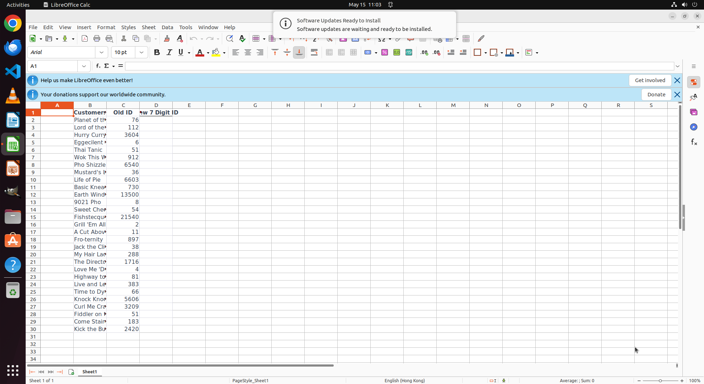}\\[-0.2ex]
{\scriptsize\ttfamily A. initial state}
\end{minipage}\hfill
\begin{minipage}[t]{0.48\linewidth}
\centering
\includegraphics[width=\linewidth]{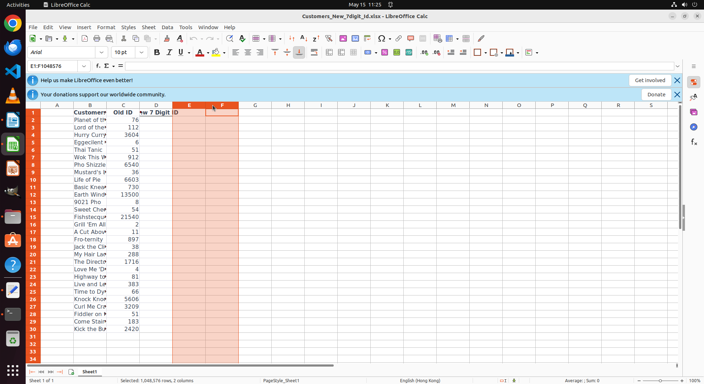}\\[-0.2ex]
{\scriptsize\ttfamily B. attempted end state}
\end{minipage}
\end{rainputbox}

\begin{ratoolbox}
\raStepHeader{01}{execute\_vm\_command}
\raCmd{\$ find /home -maxdepth 4 -type f | sed 's\#\textasciicircum{}\#/\#\#' | head -200}
\raThought{\# thought: I need to verify the actual spreadsheet contents, not just the visible screenshot, because padding with}
\raThought{leading zeros may be represented as text/formatting and many rows are involved. The active document title indicates a}
\raThought{file named "Customers\_New\_7digit\_id.xlsx", so I will inspect the filesystem to locate the spreadsheet and then read}
\raThought{its cell values.}
\raObs{=> //home/user/.gnupg/pubring.kbx //home/user/.gnupg/trustdb.gpg //home/user/.bashrc //home/user/server/main.py}
\raObs{//home/user/server/screenshots/screenshot.png //home/user/server/out.log...}
\smallskip
\raStepHeader{02}{execute\_vm\_command}
\raCmd{\$ python3 - <<'PY' from openpyxl import load\_workbook path='/home/user/Customers\_New\_7digit\_Id.xlsx'}
\raCmd{wb=load\_workbook(path,data\_only=False) ws=wb.active print('sheet', ws.title, 'max\_row', ws.max\_row, 'max\_col',}
\raCmd{ws.max\_column) for r in range(1, min(ws.max\_row,40)+1): vals=[] for c in range(1, min(ws.max\_column,8)+1):}
\raCmd{cell=ws.cell(r,c) vals.append((cell.coordinate, cell.value, cell.number\_format, cell.data\_type)) print(vals) PY}
\raThought{\# thought: The file is present at `/home/user/Customers\_New\_7digit\_Id.xlsx`. I need to inspect the workbook cells to}
\raThought{confirm whether the "New 7 Digit Id" column contains the corresponding Old ID values padded to seven digits, and}
\raThought{whether unrelated columns were changed. I will read the workbook values and formats with Python.}
\raObs{=> sheet Sheet1 max\_row 32 max\_col 4 [('A1', None, 'General', 'n'), ('B1', 'Customers', 'General', 's'), ('C1', 'Old}
\raObs{ID', 'General', 's'), ('D1', 'New 7 Digit ID', 'General', 's')] [('A2',...}
\smallskip
\end{ratoolbox}

\begin{rafinalbox}
\raStepHeader{03}{final\_answer}
\raCmd{reward=1 verdict=Success}
\raThought{\# thought: The expected final state is that every value in the existing "Old ID" column (C2:C30) is copied into the}
\raThought{corresponding row of "New 7 Digit ID" (D2:D30) as a seven-character identifier with leading zeros, while other}
\raThought{customer data and irrelevant blank regions remain unchanged/empty. Visual evidence shows D was populated, and}
\raThought{file-backed verification confirms each D value is exactly the zero-padded form of the corresponding C value, with no}
\raThought{data added beyond column D. Therefore all critical criteria are satisfied.}
\raObs{=> Evaluation completed}
\raFinalText{The workbook `/home/user/Customers\_New\_7digit\_Id.xlsx` was inspected directly. Rows D2:D30 under "New 7 Digit ID"}
\raFinalText{contain the exact seven-character zero-padded versions of the corresponding Old ID values in C2:C30, e.g. 76 ->}
\raFinalText{0000076, 3604 -> 0003604, 13500 -> 0013500, and all remaining rows similarly match. The customer names and old IDs}
\raFinalText{remain present, and the workbook has no data beyond the intended columns; irrelevant blank regions are empty. The}
\raFinalText{final state satisfies the task fully.}
\end{rafinalbox}

%% file: supplementary_sections/error_case_analysis.tex
\subsection{Interactive Reward Agent Error Case Analysis}
\label{sec:appendix_error_cases}

\begin{figure}[h]
    \centering
    \includegraphics[width=\linewidth]{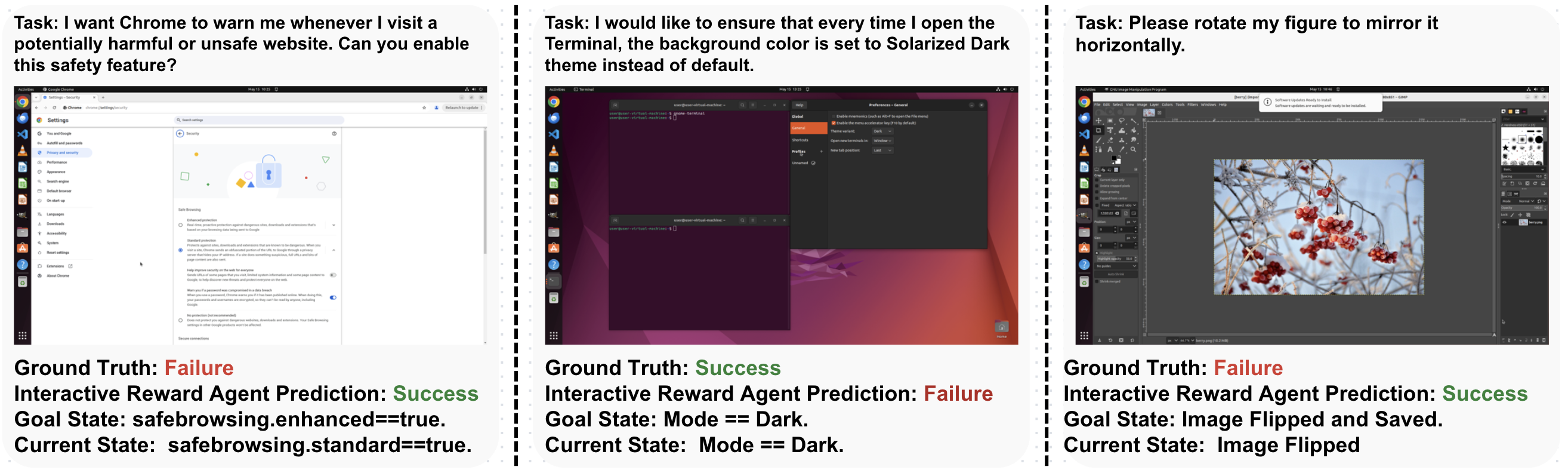}
    \caption{Representative IRA failures caused by misaligned completion conditions: incorrect protection granularity, an overly literal theme interpretation, and failure to verify that an edited artifact was saved.}
    \label{fig:failurecase}
\end{figure}

Fig.~\ref{fig:failurecase} demonstrates failures caused by misaligned completion
conditions rather than insufficient evidence acquisition. IRA
incorrectly accepts standard protection for a task requiring
enhanced protection, rejects a valid dark-mode state by interpreting “Solarized Dark” too literally, and infers completion from a historical horizontal-flip operation without verifying the implicit requirement that the flipped image be saved. These cases show that user instructions may be ambiguous or omit expectations such as acceptable semantic equivalents and persistence. Consequently, IRA may formulate incorrect completion conditions even when relevant evidence is available. Robust evaluation therefore requires better alignment between the literal instruction, the user’s underlying intent, and the completion conditions used for verification. More cases and detailed trajectories are provided in the following.

\noindent\textbf{False positive: operation evidence without saved artifact}
Please rotate my figure to mirror it horizontally.

\vspace{1mm}
\ecBadge{case: gimp\_mirror\_unsaved} \ecBadge{model: gpt-5.5} \ecBadge{steps: 6} \ecBadge{category: Implicit completion requirement} \ecGoodBadge{Reward Agent: Success (1.00)} \ecBadBadge{GT:  (0.00)}

\begin{ecinputbox}
\ecBoxTitle{Screenshots provided to the reward model}
\ecThought{VLM can see the start/end desktop state, but not hidden file, config, post-execution configuration state, or application state.}
\vspace{1mm}
\begin{minipage}[t]{0.31\linewidth}
  \centering
  \includegraphics[width=\linewidth,height=0.19\textheight,keepaspectratio]{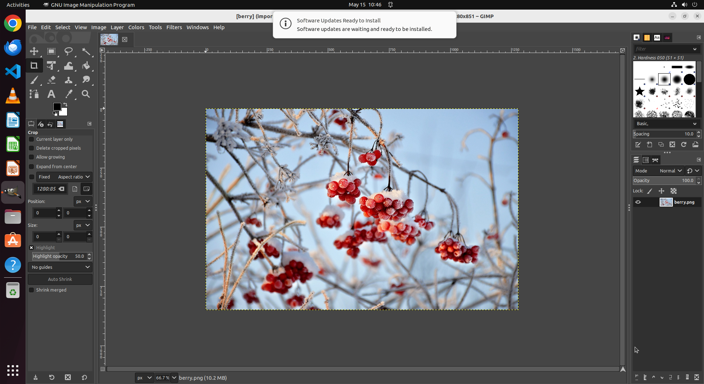}\\[-1mm]
  {\ttfamily\tiny A. Initial image state}
\end{minipage}%
\hfill%
\begin{minipage}[t]{0.31\linewidth}
  \centering
  \includegraphics[width=\linewidth,height=0.19\textheight,keepaspectratio]{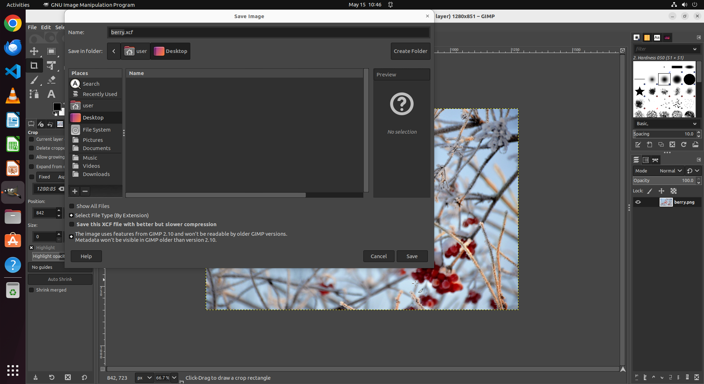}\\[-1mm]
  {\ttfamily\tiny B. Attempted end state: save dialog still visible}
\end{minipage}%
\hfill%
\begin{minipage}[t]{0.31\linewidth}
  \centering
  \includegraphics[width=\linewidth,height=0.19\textheight,keepaspectratio]{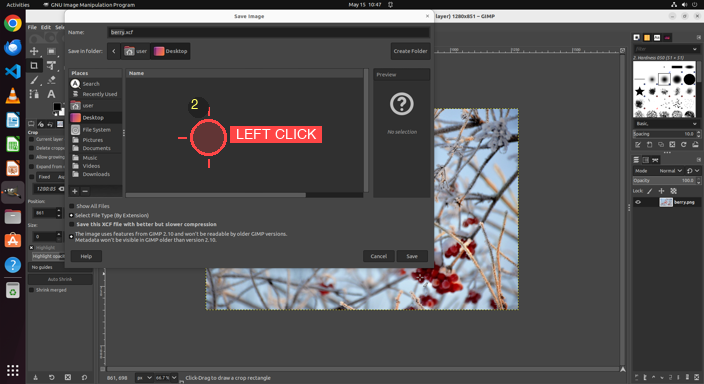}\\[-1mm]
  {\ttfamily\tiny C. Reward Agent GUI verification attempt}
\end{minipage}
\end{ecinputbox}

\begin{ectoolbox}
\ecStepHeader{Task reading}{What the task should require}
\ecFinalText{Mirror the figure horizontally and leave the edited figure saved/exported as the final artifact. An undo-history entry alone is not enough.}
\ecStepHeader{Error mechanism}{Why the Reward Agent judgment is wrong}
\ecFinalText{Reward Agent relied on GIMP state evidence such as 'Undo Flip Layer' and treated the open save dialog as irrelevant. The failure is task-semantics calibration: the model did not infer the persistence requirement hidden in the user's editing request.}
\ecStepHeader{Key verification steps}{Evidence used by the judge}
\begin{enumerate}[leftmargin=*,nosep]
\item \textbf{Step 1-3}: Reward Agent tried to dismiss the save dialog to inspect the visible canvas.
\item \textbf{Step 4}: It inspected the accessibility tree and found evidence of a flip operation in the undo history.
\item \textbf{Final}: It returned Success because the visual operation occurred, explicitly saying no save/export requirement was specified.
\end{enumerate}
\end{ectoolbox}

\begin{ecfinalbox}
\ecStepHeader{Final judgment}{Reward Agent output and paper takeaway}
\ecFinalText{Reward Agent returned Success with reward 1.00. The model verified that the operation happened, but missed that an image-editing request usually requires a persisted output.}
\end{ecfinalbox}

\noindent\textbf{False negative: transient controls mistaken for non-fullscreen state}
Can you enable fullscreen mode in VLC so that the video fills up the whole screen?

\vspace{1mm}
\ecBadge{case: vlc\_fullscreen\_overlay} \ecBadge{model: gpt-5.5} \ecBadge{steps: 2} \ecBadge{category: Visual overconfidence} \ecBadBadge{Reward Agent: Failure (0.00)} \ecGoodBadge{GT: True (1.00)}

\begin{ecinputbox}
\ecBoxTitle{Screenshots provided to the reward model}
\ecThought{VLM can see the start/end desktop state, but not hidden file, config, post-execution configuration state, or application state.}
\vspace{1mm}
\begin{minipage}[t]{0.47\linewidth}
  \centering
  \includegraphics[width=\linewidth,height=0.19\textheight,keepaspectratio]{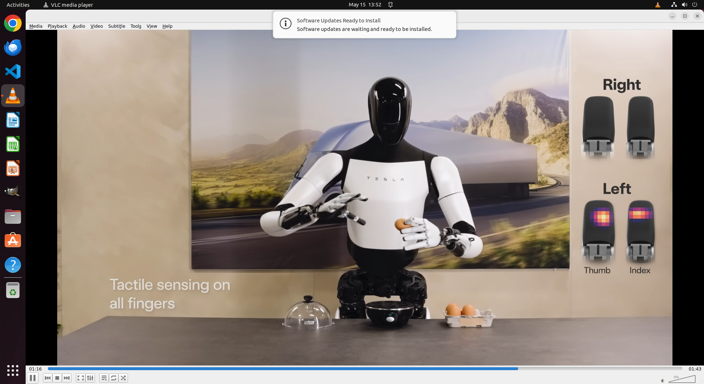}\\[-1mm]
  {\ttfamily\tiny A. Initial VLC state}
\end{minipage}%
\hfill%
\begin{minipage}[t]{0.47\linewidth}
  \centering
  \includegraphics[width=\linewidth,height=0.19\textheight,keepaspectratio]{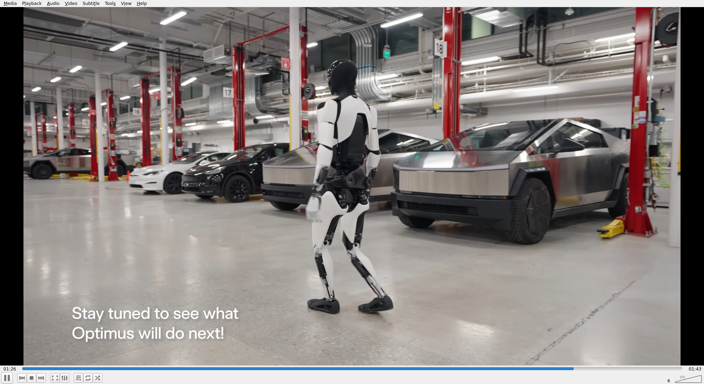}\\[-1mm]
  {\ttfamily\tiny B. Attempted end state with visible overlay}
\end{minipage}
\end{ecinputbox}

\begin{ectoolbox}
\ecStepHeader{Task reading}{What the task should require}
\ecFinalText{Enable VLC fullscreen. Temporary playback controls or progress overlays do not necessarily imply the video is windowed.}
\ecStepHeader{Error mechanism}{Why the Reward Agent judgment is wrong}
\ecFinalText{Reward Agent stopped after comparing screenshots. It treated the visible menu/progress controls as decisive evidence of non-fullscreen, although the benchmark oracle checked VLC window size against screen size.}
\ecStepHeader{Key verification steps}{Evidence used by the judge}
\begin{enumerate}[leftmargin=*,nosep]
\item \textbf{Final only}: Reward Agent made the failure decision directly from the screenshot.
\item \textbf{Missed check}: It did not query window size or application fullscreen state.
\end{enumerate}
\end{ectoolbox}

\begin{ecfinalbox}
\ecStepHeader{Final judgment}{Reward Agent output and paper takeaway}
\ecFinalText{Reward Agent returned Failure with reward 0.00. The model over-trusted visible controls and did not verify the actual window/screen geometry.}
\end{ecfinalbox}

\noindent\textbf{False positive: correct setting family but wrong protection tier}
I want Chrome to warn me whenever I visit a potentially harmful or unsafe website. Can you enable this safety feature?

\vspace{1mm}
\ecBadge{case: chrome\_standard\_vs\_enhanced} \ecBadge{model: gpt-5.5} \ecBadge{steps: 6} \ecBadge{category: Wrong task granularity} \ecGoodBadge{Reward Agent: Success (1.00)} \ecBadBadge{GT:  (0.00)}

\begin{ecinputbox}
\ecBoxTitle{Screenshots provided to the reward model}
\ecThought{VLM can see the start/end desktop state, but not hidden file, config, post-execution configuration state, or application state.}
\vspace{1mm}
\begin{minipage}[t]{0.47\linewidth}
  \centering
  \includegraphics[width=\linewidth,height=0.19\textheight,keepaspectratio]{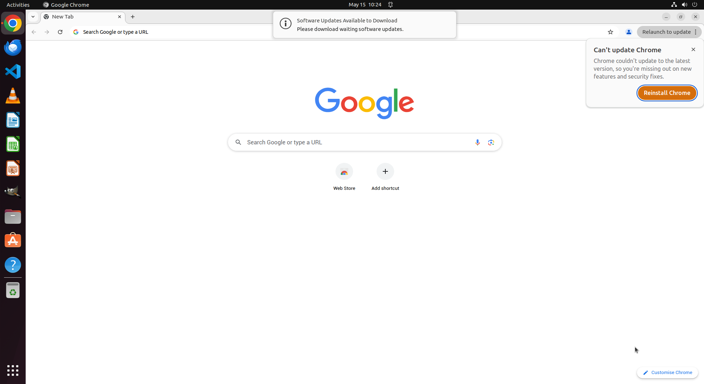}\\[-1mm]
  {\ttfamily\tiny A. Initial Chrome state}
\end{minipage}%
\hfill%
\begin{minipage}[t]{0.47\linewidth}
  \centering
  \includegraphics[width=\linewidth,height=0.19\textheight,keepaspectratio]{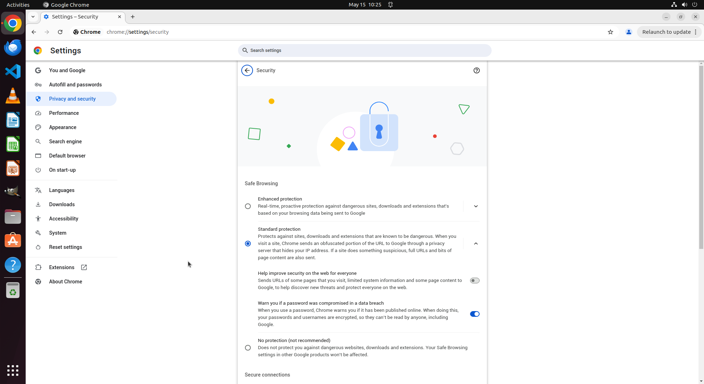}\\[-1mm]
  {\ttfamily\tiny B. Security settings inspected by Reward Agent}
\end{minipage}
\end{ecinputbox}

\begin{ectoolbox}
\ecStepHeader{Task reading}{What the task should require}
\ecFinalText{The target criterion is Enhanced Safe Browsing, i.e., `safebrowsing.enhanced == true`; Standard Safe Browsing is insufficient.}
\ecStepHeader{Error mechanism}{Why the Reward Agent judgment is wrong}
\ecFinalText{Reward Agent correctly navigated to Chrome security settings, but collapsed Standard and Enhanced protection into the same semantic bucket. This is not a tool-access failure; it is a criterion-level interpretation error.}
\ecStepHeader{Key verification steps}{Evidence used by the judge}
\begin{enumerate}[leftmargin=*,nosep]
\item \textbf{Step 1}: It inspected Chrome preference files for Safe Browsing state.
\item \textbf{Step 2-4}: It opened `chrome://settings/security` for direct UI evidence.
\item \textbf{Final}: It accepted Standard protection as satisfying the safety-warning request.
\end{enumerate}
\end{ectoolbox}

\begin{ecfinalbox}
\ecStepHeader{Final judgment}{Reward Agent output and paper takeaway}
\ecFinalText{Reward Agent returned Success with reward 1.00. The model matched the broad concept of safety browsing but not the exact required tier.}
\end{ecfinalbox}

\noindent\textbf{False negative: exact theme identity over a functional dark-theme criterion}
I would like to ensure that every time I open the Terminal, the background color is set to Solarized Dark theme instead of default. Please guide me to set it permanently.

\vspace{1mm}
\ecBadge{case: terminal\_solarized\_dark} \ecBadge{model: gpt-5.5} \ecBadge{steps: 7} \ecBadge{category: Over-literal interpretation} \ecBadBadge{Reward Agent: Failure (0.00)} \ecGoodBadge{GT: True (1.00)}

\begin{ecinputbox}
\ecBoxTitle{Screenshots provided to the reward model}
\ecThought{VLM can see the start/end desktop state, but not hidden file, config, post-execution configuration state, or application state.}
\vspace{1mm}
\begin{minipage}[t]{0.47\linewidth}
  \centering
  \includegraphics[width=\linewidth,height=0.19\textheight,keepaspectratio]{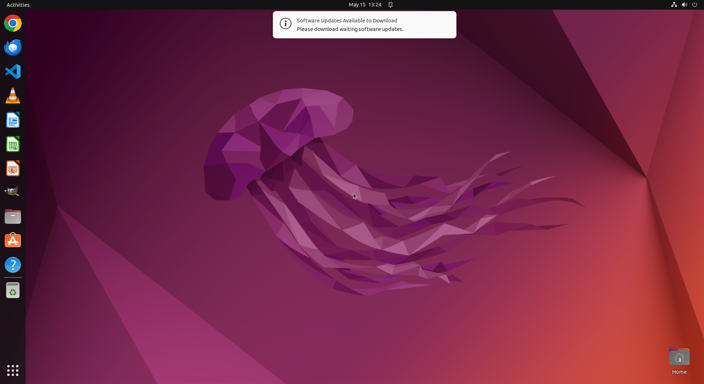}\\[-1mm]
  {\ttfamily\tiny A. Initial terminal preferences state}
\end{minipage}%
\hfill%
\begin{minipage}[t]{0.47\linewidth}
  \centering
  \includegraphics[width=\linewidth,height=0.19\textheight,keepaspectratio]{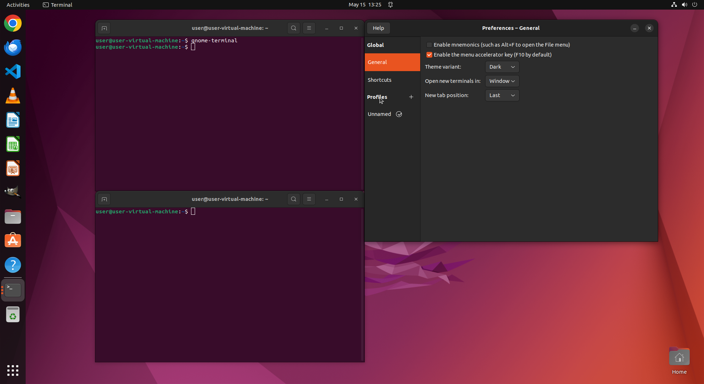}\\[-1mm]
  {\ttfamily\tiny B. Attempted end state}
\end{minipage}
\end{ecinputbox}

\begin{ectoolbox}
\ecStepHeader{Task reading}{What the task should require}
\ecFinalText{For this task family, the accepted success condition is that Terminal opens in a dark theme permanently; exact Solarized palette matching is not required by the oracle.}
\ecStepHeader{Error mechanism}{Why the Reward Agent judgment is wrong}
\ecFinalText{Reward Agent correctly inspected gsettings and found the app theme was dark but the profile palette was not exactly Solarized Dark. The failure is over-specific interpretation of a flexible user request.}
\ecStepHeader{Key verification steps}{Evidence used by the judge}
\begin{enumerate}[leftmargin=*,nosep]
\item \textbf{Step 1-3}: It tried dconf dumps and treated empty output as inconclusive.
\item \textbf{Step 4-5}: It used gsettings to inspect the profile and theme settings.
\item \textbf{Final}: It returned Failure because the profile was not exact Solarized Dark.
\end{enumerate}
\end{ectoolbox}

\begin{ecfinalbox}
\ecStepHeader{Final judgment}{Reward Agent output and paper takeaway}
\ecFinalText{Reward Agent returned Failure with reward 0.00. The model enforced a stricter theme identity than the benchmark's effective requirement.}
\end{ecfinalbox}

\noindent\textbf{Thresholded partial: title formatting correct but scope over-expanded}
Bold the text on slide 1. Make the title of size 44pt and underline it on slide 1. The file to be evaluated is located on the VM at the following path: "/home/user/Desktop/39\_2.pptx".

\vspace{1mm}
\ecBadge{case: ppt\_bold\_title\_scope} \ecBadge{model: gpt-5.5} \ecBadge{steps: 3} \ecBadge{category: Scope over-expansion} \ecBadBadge{Reward Agent: Partial Success (0.67)} \ecGoodBadge{GT: True (1.00)}

\begin{ecinputbox}
\ecBoxTitle{Screenshots provided to the reward model}
\ecThought{VLM can see the start/end desktop state, but not hidden file, config, post-execution configuration state, or application state.}
\vspace{1mm}
\begin{minipage}[t]{0.47\linewidth}
  \centering
  \includegraphics[width=\linewidth,height=0.19\textheight,keepaspectratio]{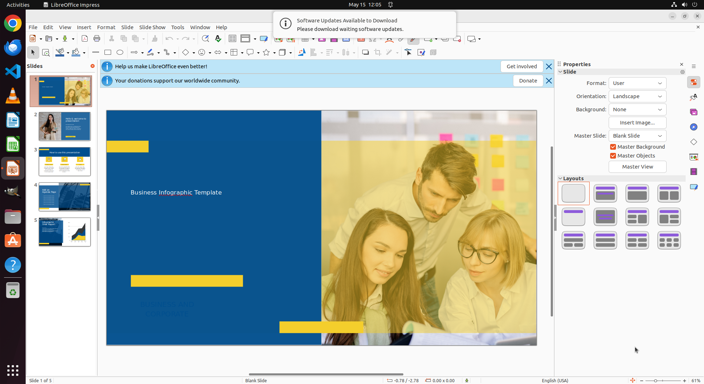}\\[-1mm]
  {\ttfamily\tiny A. Initial Impress state}
\end{minipage}%
\hfill%
\begin{minipage}[t]{0.47\linewidth}
  \centering
  \includegraphics[width=\linewidth,height=0.19\textheight,keepaspectratio]{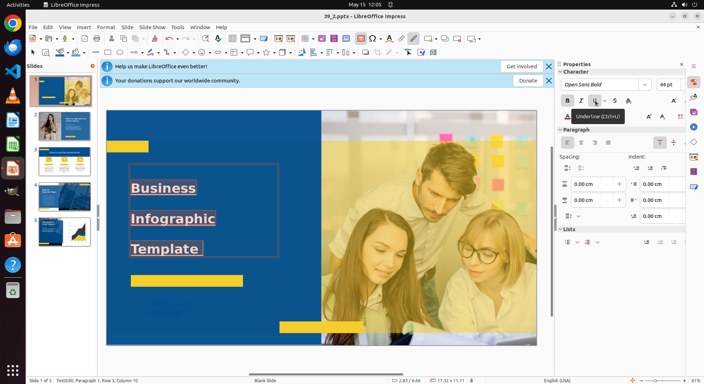}\\[-1mm]
  {\ttfamily\tiny B. Formatted slide state}
\end{minipage}
\end{ecinputbox}

\begin{ectoolbox}
\ecStepHeader{Task reading}{What the task should require}
\ecFinalText{The likely intended focus is the slide title formatting plus visible slide text; the oracle accepts the edited PPTX as matching the gold file.}
\ecStepHeader{Error mechanism}{Why the Reward Agent judgment is wrong}
\ecFinalText{Reward Agent inspected PPTX XML and correctly found the title bold/44pt/underlined, but also found another text object not bold. The issue is scope expansion in criterion parsing.}
\ecStepHeader{Key verification steps}{Evidence used by the judge}
\begin{enumerate}[leftmargin=*,nosep]
\item \textbf{Step 1}: It opened slide XML for file-backed formatting evidence.
\item \textbf{Final}: It gave 0.667 because one non-title text object was not bold.
\end{enumerate}
\end{ectoolbox}

\begin{ecfinalbox}
\ecStepHeader{Final judgment}{Reward Agent output and paper takeaway}
\ecFinalText{Reward Agent returned Partial Success with reward 0.67. The model interpreted 'text on slide 1' more broadly than the evaluator and penalized an extra text box.}
\end{ecfinalbox}

\noindent\textbf{Thresholded partial: correct image and size, strict coordinate criterion}
Insert the image 'none.png' from the Desktop onto slide 2 at the top-left corner with size 4cm*3cm.

\vspace{1mm}
\ecBadge{case: ppt\_image\_position\_strict} \ecBadge{model: gpt-5.5} \ecBadge{steps: 5} \ecBadge{category: Over-strict spatial criterion} \ecBadBadge{Reward Agent: Partial Success (0.67)} \ecGoodBadge{GT: True (1.00)}

\begin{ecinputbox}
\ecBoxTitle{Screenshots provided to the reward model}
\ecThought{VLM can see the start/end desktop state, but not hidden file, config, post-execution configuration state, or application state.}
\vspace{1mm}
\begin{minipage}[t]{0.47\linewidth}
  \centering
  \includegraphics[width=\linewidth,height=0.19\textheight,keepaspectratio]{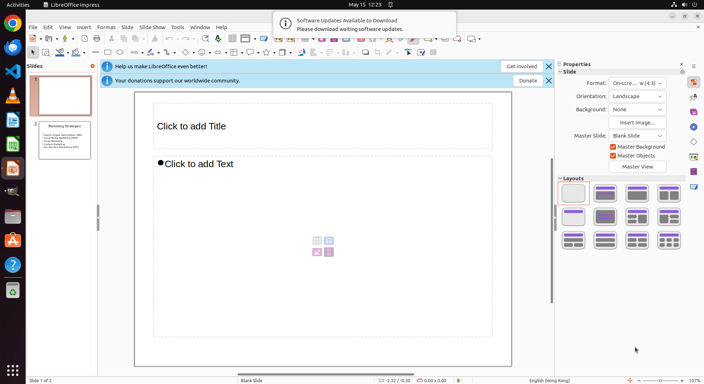}\\[-1mm]
  {\ttfamily\tiny A. Initial slide state}
\end{minipage}%
\hfill%
\begin{minipage}[t]{0.47\linewidth}
  \centering
  \includegraphics[width=\linewidth,height=0.19\textheight,keepaspectratio]{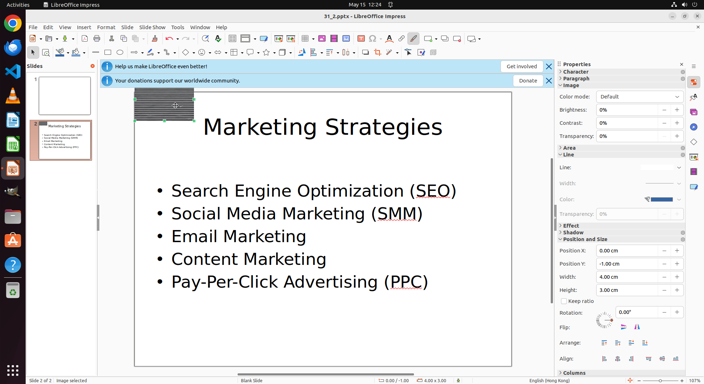}\\[-1mm]
  {\ttfamily\tiny B. Inserted image selected in Impress}
\end{minipage}
\end{ecinputbox}

\begin{ectoolbox}
\ecStepHeader{Task reading}{What the task should require}
\ecFinalText{Insert the requested image on slide 2 at top-left with 4 cm by 3 cm size. Depending on UI coordinate conventions, a near-edge placement may still satisfy the practical goal.}
\ecStepHeader{Error mechanism}{Why the Reward Agent judgment is wrong}
\ecFinalText{Reward Agent compared embedded-image hash and EMU size successfully, then penalized `y=-1cm`. This is a strict spatial interpretation rather than failure to verify the artifact.}
\ecStepHeader{Key verification steps}{Evidence used by the judge}
\begin{enumerate}[leftmargin=*,nosep]
\item \textbf{Step 1}: It located the PPTX and source image files.
\item \textbf{Step 2-3}: It inspected slide XML and compared image hashes.
\item \textbf{Final}: It returned 0.667 because the vertical offset was -1 cm.
\end{enumerate}
\end{ectoolbox}

\begin{ecfinalbox}
\ecStepHeader{Final judgment}{Reward Agent output and paper takeaway}
\ecFinalText{Reward Agent returned Partial Success with reward 0.67. The model validated the image and size exactly but treated a small coordinate offset as decisive.}
\end{ecfinalbox}